\documentclass[10pt,twocolumn,letterpaper]{article}

\usepackage[pagenumbers]{cvpr} 

\usepackage{graphicx}
\usepackage{amsmath}
\usepackage{amssymb}
\usepackage{booktabs}
\usepackage{times}
\usepackage{epsfig}
\usepackage{enumitem}

\usepackage{multirow}
\usepackage{soul}
\usepackage{pifont} 
\usepackage[T1]{fontenc}
\usepackage{mathabx,graphicx}
\usepackage{makecell}
\usepackage[dvipsnames]{xcolor}
\usepackage{color, colortbl}
\usepackage{caption}

\def\CircleArrowright{\ensuremath{%
  \rotatebox[origin=c]{310}{$\circlearrowright$}}}

\newcommand\Tstrut{\rule{0pt}{2.3ex}}         
\newcommand{\vlnbert}{VLN$\protect\CircleArrowright$BERT}
\newcommand{\high}[1]{{\textbf{\color{blue}#1}}}

\usepackage[pagebackref,breaklinks,colorlinks]{hyperref}

\def\cvprPaperID{1363} 
\def\confName{CVPR}
\def\confYear{2022}

\begin{document}

\title{Bridging the Gap Between Learning in Discrete and Continuous\\ Environments for Vision-and-Language Navigation}

\author{
Yicong Hong$^{1*}$ \quad Zun Wang$^{1*}$ \quad Qi Wu$^2$ \quad Stephen Gould$^1$\\
$^1$The Australian National University, $^2$University of Adelaide\\
{\tt\small \{yicong.hong, zun.wang, stephen.gould\}@anu.edu.au} \\
{\tt\small qi.wu01@adelaide.edu.au} \\ \\
{\tt\small Project URL: \url{https://github.com/YicongHong/Discrete-Continuous-VLN}}
}

\vspace{-1pt}

\twocolumn[{
\renewcommand\twocolumn[1][]{#1}
\maketitle
\begin{center}
    \centering
    \captionsetup{type=figure}
    \includegraphics[width=\textwidth]{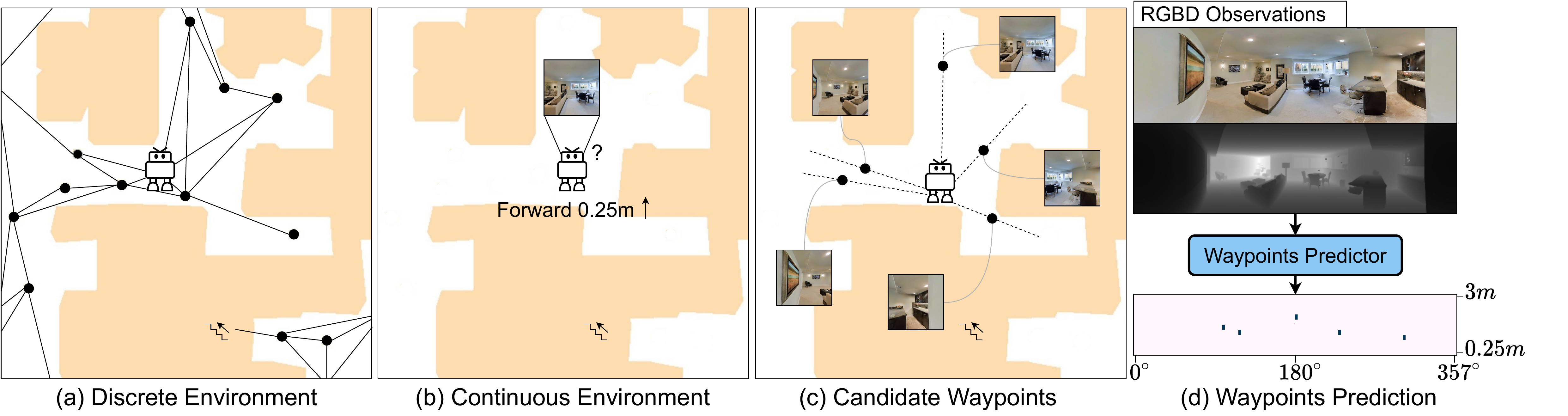}
    \captionof{figure}{Vision-and-language navigation in discrete versus in continuous environments. (a) Agents in discrete environments rely on the connectivity graph to navigate with panoramic high-level actions, (b) but they need to perform low-level controls to move in continuous spaces. (c,d) We propose a candidate waypoints predictor to predict accessible positions in continuous environments, to bridge the discrete-to-continuous gap.}
    \label{fig:intro}
\end{center}
}]

\begin{abstract}
    Most existing works in vision-and-language navigation~(VLN) focus on either discrete or continuous environments, training agents that cannot generalize across the two. Although learning to navigate in continuous spaces is closer to the real-world, training such an agent is significantly more difficult than training an agent in discrete spaces.
    However, recent advances in discrete VLN are challenging to translate to continuous VLN due to the domain gap.
    The fundamental difference between the two setups is that discrete navigation assumes prior knowledge of the connectivity graph of the environment, so that the agent can effectively transfer the problem of navigation with low-level controls to jumping from node to node with high-level actions by grounding to an image of a navigable direction. 

    To bridge the discrete-to-continuous gap, we propose a
    predictor to generate a set of candidate waypoints during navigation, so that agents designed with high-level actions can be transferred to and trained in continuous environments. We refine the connectivity graph of Matterport3D to fit the continuous Habitat-Matterport3D, and train the waypoints predictor with the refined graphs to produce accessible waypoints at each time step. Moreover, we demonstrate that the predicted waypoints can be augmented during training to diversify the views and paths, and therefore enhance agent's generalization ability.

    Through extensive experiments we show that agents navigating in continuous environments with predicted waypoints perform significantly better than agents using low-level actions, which reduces the absolute discrete-to-continuous gap by 11.76\% Success Weighted by Path Length (SPL) for the Cross-Modal Matching Agent and 18.24\% SPL for the \vlnbert. Our agents, trained with a simple imitation learning objective, outperform previous methods by a large margin, achieving new state-of-the-art results on the testing environments of the R2R-CE and the RxR-CE datasets.
    {\let\thefootnote\relax\footnote{{* Authors contributed equally}}}
\end{abstract}

\vspace{-5pt}


\section{Introduction}
\label{sec:introduction}

Vision-and-language navigation (VLN)~\cite{anderson2018vision} is a challenging cross-domain research problem which requires an agent to interpret human instructions and navigate in previously unseen environments by executing a sequence of actions. Two distinct scenarios have been proposed for VLN research, being navigation in discrete environments (R2R, RxR~\cite{anderson2018vision,anderson2020rxr}) and in continuous environments (R2R-CE, RxR-CE~\cite{krantz2020navgraph}).
Due to the large domain gap, navigation in the two scenarios have been studied independently in previous works, and a large number of prominent advances achieved by agents in discrete spaces, such as by leveraging pre-trained visiolinguistic transformers~\cite{hao2020towards,hong2020recurrent} or by using scene memory~\cite{deng2020evolving,wang2021structured}
cannot be directly applied to agents traversing continuous spaces. 

The fundamental difference between navigation in discrete and in continuous environments is the reliance on the connectivity graph which contains numbers of sparse nodes (waypoints) distributed across accessible spaces of the environment. With the prior knowledge of the connectivity graph, the agent can move with a panoramic high-level action space~\cite{fried2018speaker}, \ie, teleport to an adjacent waypoint on the graph by selecting a single direction from the discrete set of navigable directions. Compared to navigation in continuous environments, which usually relies on a limited field of view to infer low-level controls (\eg, turn left 15 degrees or move forward 0.25 meters)~\cite{krantz2020navgraph}, navigation with panoramic actions and the connectivity graph simplifies the complicated decision making problem by formulating it as an explicit text-to-image grounding task. First, the agents are not required to infer the important concepts of accessibility (openspace vs. obstacles) from sensory inputs. Second, distinct visual representations can be defined for each navigable direction, so the agents only need to match contextual clues from instruction to visual options to move, which greatly reduces the agent's state space and facilitates the learning.
As a result, many previous works in VLN with high-level actions address the navigation problem mostly from the visual-textual matching perspective. A large number of innovations such as back-translation~\cite{tan2019learning}, back-tracking~\cite{ke2019tactical,ma2019regretful}, scene memory~\cite{deng2020evolving,wang2021structured}
and transformer-based pre-training~\cite{hao2020towards,majumdar2020improving,hong2020recurrent} bring remarkable improvement but they cannot be directly transferred to agents in continuous environments. There remains about a 20\% gap in success rate for agents with the same architecture navigating in discrete and in continuous spaces~\cite{krantz2020navgraph}. 

Despite the great efficiency of learning in a discrete environment, navigation in continuous spaces is much closer to the real-world. In this paper, we address the problem of bridging the learning between the two domains, aiming to effectively adapt agents designed for discrete VLN to continuous environments. First of all, we identify and quantitatively evaluate the value of high-level controls in VLN, showing the importance of knowing accessible waypoints. Second, inspired by Sim2Real-VLN~\cite{anderson2020sim}, we introduce a powerful candidate waypoints predictor to estimate navigable locations in continuous spaces, the module constructs a local navigability graph centered at the agent at each time step using the visual observations. In Sim2Real-VLN~\cite{anderson2020sim}, the sub-goal module is trained on the pre-defined connectivity graphs of the Matterport3D environment (MP3D)~\cite{chang2017matterport3d}, where there exist edges going through obstacles and nodes in inaccessible spaces. In contrast, we transfer the discrete MP3D graph onto the continuous Habitat-MP3D~\cite{savva2019habitat} spaces, and represent the transferred waypoints as targets for learning a mixture of Gaussian probability map, resulting in a robust waypoints predictor in unvisited environments that supports navigation. Furthermore, we propose a simple augmentation method to move the position of waypoints during training of the agent, so that the agent can learn to reach the same target using diverse observations and step lengths, thereby improving the generalization ability.

With the proposed candidate waypoints predictor, we evaluate the performance of agents designed for discrete VLN in continuous environments. The selected agents, including the cross-modal matching agent (CMA)~\cite{wang2019reinforced} and the \vlnbert~\cite{hong2020recurrent} are widely applied methods that are distinct in network architecture. Our experiments show that agents in continuous environments trained with the predicted waypoints significantly improves over navigation without using waypoints, reducing the discrete-continuous gap, and achieving new state-of-the-art performance of 39\% and 19.61\% SPL on the benchmarking R2R-CE and RxR-CE test sets~\cite{anderson2018vision,krantz2020navgraph,anderson2020rxr}, respectively. This results suggest that our proposed candidate waypoints predictor can enable an effective discrete-to-continuous transfer as well as showing a huge potential of benefiting other navigation problems.

\section{Related Work} 
\label{sec:relatedwork}

\paragraph{Vision-and-Language Navigation}
Visiolinguistic cross-modal grounding is a crucial skill for addressing vision-and-language navigation problems. Tasks for indoor navigation with low-level instructions such as R2R~\cite{anderson2018vision} and RxR~\cite{anderson2020rxr}, outdoor navigation such as Touchdown~\cite{chen2019touchdown}, navigation with dialog such as CVDN~\cite{thomason2020vision} and HANNA~\cite{nguyen2019help}, and navigation for remote object grounding such as REVERIE~\cite{qi2020reverie} and SOON~\cite{zhu2021soon} all require the agent's ability to associate time-dependent visual observations to instructions for decision making. To facilitate the learning, Fried~\etal~\cite{fried2018speaker} exploit the connectivity graphs defined for Matterport3D environments~\cite{chang2017matterport3d} and propose to navigate with panoramic actions, which allows teleportation of agent among adjacent nodes (waypoints) on the graph by choosing an image pointing towards the node. Following this idea, later works that are focusing on cross-modality learning~\cite{wang2019reinforced, ma2019self, hong2020graph}, data augmentation~\cite{tan2019learning,fu2020counterfactual,parvaneh2020counterfactual}, waypoint-tracking~\cite{ke2019tactical,ma2019regretful,deng2020evolving,wang2021structured}, and pre-training for Transformer-based models~\cite{li2019robust,hao2020towards,majumdar2020improving,hong2020recurrent} are implemented based on the connectivity graph and the high-level action space. Despite the great advances in learning the correspondence among vision, language and action, these agents are not applicable to the more practical scenario -- navigation in continuous environments, which requires the agent's ability of inferring spatial accessibility.

\paragraph{Continuous VLN} 
Continuous environments such as AI2-THOR~\cite{kolve2017ai2}, House3D~\cite{wu2018building}, CHALET~\cite{yan2018chalet}, Gibson~\cite{xia2018gibson}, Habitat~\cite{savva2019habitat} and iGibson~\cite{shen2020igibson} have been setup for embodied AI research in synthetic and photo-realistic scenes. To study vision-and-language navigation in continuous environments (VLN-CE), Krantz~\etal~\cite{krantz2020navgraph} transfer the discrete paths in R2R~\cite{anderson2018vision} (and RxR~\cite{anderson2020rxr}) dataset to continuous trajectories based on the Habitat simulator~\cite{savva2019habitat}, and Irshad~\etal propose a hierarchical model for inferring agent's linear and angular velocities in the Robo-VLN environment~\cite{irshad2021hierarchical}. In addition, methods such as applying semantic map representations~\cite{irshad2021sasra} and using language-aligned waypoints supervision~\cite{raychaudhuri2021law} are explored in VLN-CE. Experiments demonstrate a huge performance gap between agents that navigate in discrete and continuous environments.
\vspace{-10pt}

\paragraph{Hierarchical Visual Navigation}
Hierarchical visual navigation, involving the problem of mapping, planning and control, have been extensively studied in previous literatures~\cite{anderson2020sim,chaplot2020learning,chaplot2020neural,chen2020learning,chen2021topological,chen2019behavioral,chen2018learning,gupta2017cognitive,krantz2021waypoint,meng2020scaling,savinov2018semi}. Active Neural SLAM~\cite{chaplot2020learning} plans towards a long-term goal with a map and agent pose estimated from visual and sensory inputs. Chen~\etal~\cite{chen2021topological} construct topological maps for planning by sparsifying continuous paths collected in the environments, and Chen~\etal~\cite{chen2020learning} predicts a single audio-conditioned sub-goal at each step while our model estimates key positions around the agent. Recent work that is the most relevant to ours are the Waypoint Models~\cite{krantz2021waypoint} and the Sim2Real-VLN~\cite{anderson2020sim}. Waypoint Models predicts a coarse view (direction) and a sub-goal in that view to explore, where all predictions are conditioned on vision, language and agent's state. 
Whereas our method decouples the waypoints prediction and agent's decision making process, and leverages the learned navigability to provide accessible directions for the agent to act, 
without extra efforts in modifying network architectures or training methods for adapting the new waypoint predictor. Sim2Real-VLN also predicts adjacent sub-goals but the model is trained on the connectivity graphs defined in MP3D~\cite{chang2017matterport3d}, where there exist a large number of invalid edges going across obstacles. In contrast, we construct graph nodes over the Habitat-MP3D spaces, and we further study the idea of training agents directly in the continuous environments with high-level actions.

\section{Background} 
\label{sec:background}

In this section, we will first introduce the background of VLN in both discrete~\cite{anderson2018vision} and continuous environments~\cite{krantz2020navgraph}. We apply the reinforced cross-modal matching agent (CMA) proposed by Wang~\etal~\cite{wang2019reinforced} to explain the general formulation of a VLN network. Then, we will evaluate the value of high-level actions by some contrastive experiments as a proof of concept for this paper.

\subsection{Navigation Setups}
The task of vision-and-language navigation is formulated as follows: Given a natural language instruction $\boldsymbol{U}$ as a sequence of $l$ words ${\langle}w_1, w_2, \ldots ,w_l{\rangle}$, an agent initialized at a certain position $\boldsymbol{p}_1$ is asked to travel to a target position $\boldsymbol{p}_T$ in an environment $\mathcal{E}$ by executing a sequence of actions $\boldsymbol{a}$. The overall navigation can be considered as a Partially Observable Markov Decision Process (PMODP) ${\langle}\boldsymbol{p}_1,\boldsymbol{a}_1,\boldsymbol{p}_2,\boldsymbol{a}_2, \ldots,\boldsymbol{a}_{t-1},\boldsymbol{p}_t{\rangle}$, where each action $\boldsymbol{a}_{t}$ takes the agent to a new position $\boldsymbol{p}_{t+1}$ and the agent receives a new visual observation $\boldsymbol{V}_{\!t}$.\footnote{POMDPs also provide the agent with an intermediate reward at each time step, which we consider to be zero in this work.}

\paragraph{Navigate with High-Level Actions}
Navigation with high-level actions~\cite{anderson2018vision,fried2018speaker} is based on the connectivity graph $\mathcal{G}$ specified for each $\mathcal{E}$. Each connectivity graph contains a set of $j$ nodes (waypoints) distributed across the entire environment $\{\boldsymbol{g}_1,\boldsymbol{g}_2,\ldots,\boldsymbol{g}_j\}{\triangleq}\mathcal{G}$. Essentially, $\mathcal{G}$ discretizes $\mathcal{E}$, constraining the agent's position $\boldsymbol{p}_{t}\in\mathcal{G}$ at all time. The connectivity graph also provides the navigable directions, which greatly facilitates the learning. Following the panoramic action space proposed by Fried~\etal~\cite{fried2018speaker}, agent at each $\boldsymbol{p}_{t}$ receives a panorama which is composed of $n$ single-view inputs (RGB images) each pointing towards an adjacent waypoint, respectively. The set of visual features corresponding to those directions are represented as $\{\boldsymbol{v}^p_1,\boldsymbol{v}^p_2,\ldots,\boldsymbol{v}^p_k\}$. Each feature $\boldsymbol{v}^p_i$ is enhanced with a directional encoding $\boldsymbol{d}^p_i$ to indicate the relative orientation of the waypoint with respect to the agent's heading, denoted as $\boldsymbol{f}^p_i=\left[\boldsymbol{v}^p_i;\boldsymbol{d}^p_i\right]$, which has been shown influential in later works~\cite{hu2019you,hong2020graph}. At each navigation step, the agent's state $\boldsymbol{h}_t$ which keep tracks of all the past observations and decisions, will be updated by a recurrent network (LSTM~\cite{hochreiter1997long}) using the attended instructions $\boldsymbol{c}^{lang}_{t}$ and past decision $\boldsymbol{a}_{t-1}$ as
\begin{equation}
\boldsymbol{h}_t=\text{LSTM}\left(\left[\boldsymbol{c}^{lang}_{t};\boldsymbol{a}_{t-1}\right],h_{t-1}\right)
\label{eqn:state_lstm}
\end{equation}

Based on the formulation above, decision making in discrete environments is implemented as a feature matching problem. In a compact form, the action probability for selecting an adjacent waypoint can be expressed as 
\begin{equation}
p_{i}=\text{Softmax}\left(\left[\boldsymbol{h}_t;\boldsymbol{c}^{lang}_{t};\boldsymbol{c}^{rgb}_{t}\right]^{\top}\boldsymbol{W}_{\!p}\boldsymbol{f}^p_i\right)
\label{eqn:rcm_discrete_action}
\end{equation}
where $\boldsymbol{c}^{rgb}_{t}$ represents the attended images and $\boldsymbol{W}_{\!p}$ indicates learnable projections. As a result, the agent navigates with two effective mechanisms: (1) \textit{view selection}, the agent chooses a direction with a visual observation that has the highest correspondence to the agent's state. (2) \textit{waypoint teleportation}, the agent teleports directly to a neighbor waypoint that is positioned in the selected view.

\paragraph{Navigate with Low-Level Actions}
VLN in continuous environments (VLN-CE)~\cite{krantz2020navgraph} is established over the Habitat simulator~\cite{savva2019habitat}, where the agent's position $\boldsymbol{p}_t$ can be any point in the open space. Due to the absence of the connectivity graph, agents need to learn to identify accessible positions in space and avoid obstacles. As proposed in VLN-CE, the agent needs to infer low-level actions (turn left 15 degrees, turn right 15 degrees, move forward 0.25 meters or stop) from the egocentric observation. To encode a single front-view image with directional clues, spatial embeddings have been concatenated to each $j=1,...,16$ patch of the ResNet~\cite{he2016deep} output features before attentive pooling~\cite{krantz2020navgraph}, \ie $\boldsymbol{f}^{front}_j=[\boldsymbol{v}^{front}_j;\boldsymbol{d}_j]$ and $\boldsymbol{c}^{img}_t=\text{AttnPool}_{j=1,\ldots,16}(\boldsymbol{f}^{front}_j)$. The agent's state is updated also using Eq.\ref{eqn:state_lstm} but the action probability is computed as
\begin{equation}
\boldsymbol{p}=\text{Softmax}\left(\left[\boldsymbol{h}_t;\boldsymbol{c}^{lang}_{t};\boldsymbol{c}^{rgb}_{t};\boldsymbol{c}^{d}_{t}\right]\boldsymbol{W}_{\!c}\right)
\label{eqn:rcm_continuous_action}
\end{equation}
where $\boldsymbol{c}^{d}_{t}$ represents the attended depth features and $\boldsymbol{W}_{\!c}$ projects the concatenated encodings to $\mathbb{R}^{4}$, corresponding to the four pre-defined actions.

With the above formulation, the learning of language-visual and language-directional correspondences becomes much more implicit. The navigation episodes with low-level actions is about 10 times longer than high-level actions, which leads to a very expensive training process.

\paragraph{Evaluation Metrics}
There are five standard metrics in VLN for evaluating the agent's performance, including Trajectory Length (TL), Navigation Error (NE), Success Rate (SR), normalized inverse of the Path Length (SPL) \cite{anderson2018evaluation}, normalized Dynamic Time Warping (nDTW) and Success weighted by normalized Dynamic Time Warping (SDTW) \cite{ilharco2019general}. See Appendix for more details.

\subsection{What is the Value of High-Level Actions?}

\begin{figure}[t]
  \centering
  \includegraphics[width=\columnwidth]{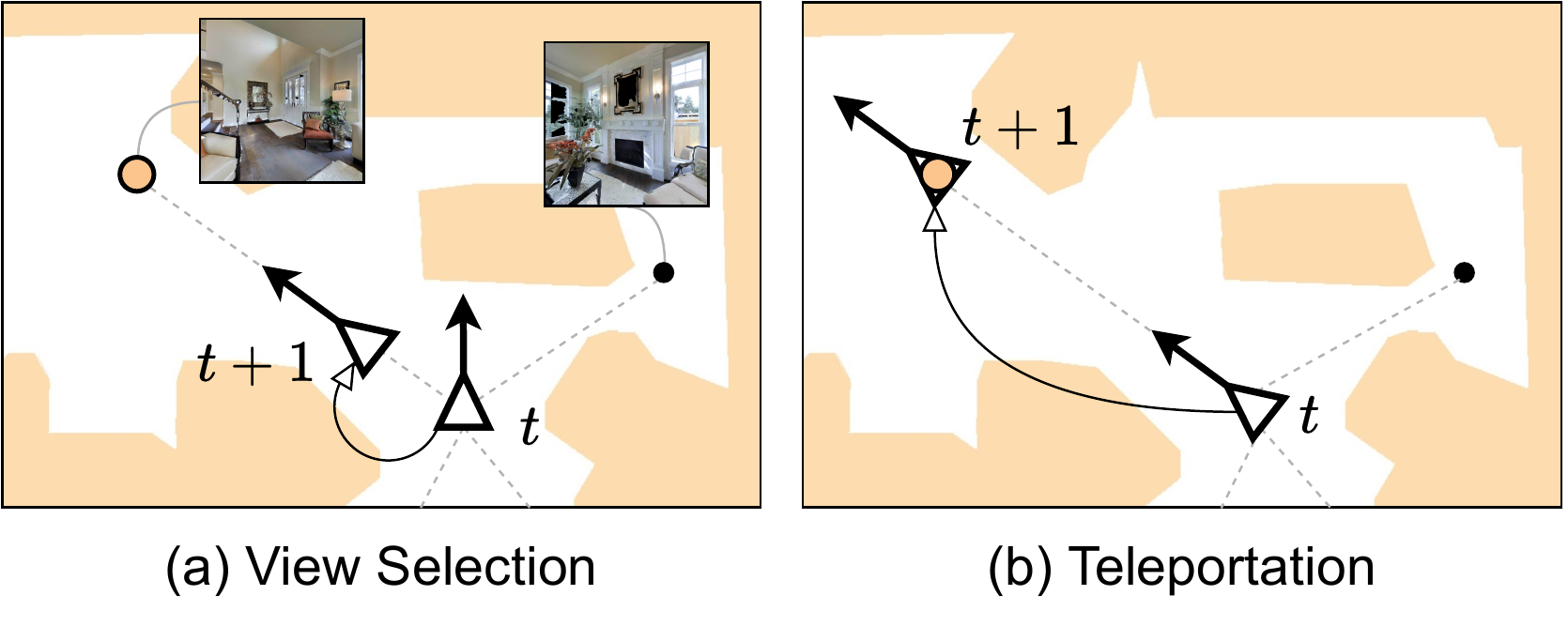}
  \vspace{-15pt}
  \caption{Navigation by (a) view selection and (b) waypoint teleportation in continuous environments.}
  \label{fig:select_teleport}
  \vspace{-5pt}
\end{figure}

\begin{figure}[t]
  \centering
  \includegraphics[width=\columnwidth]{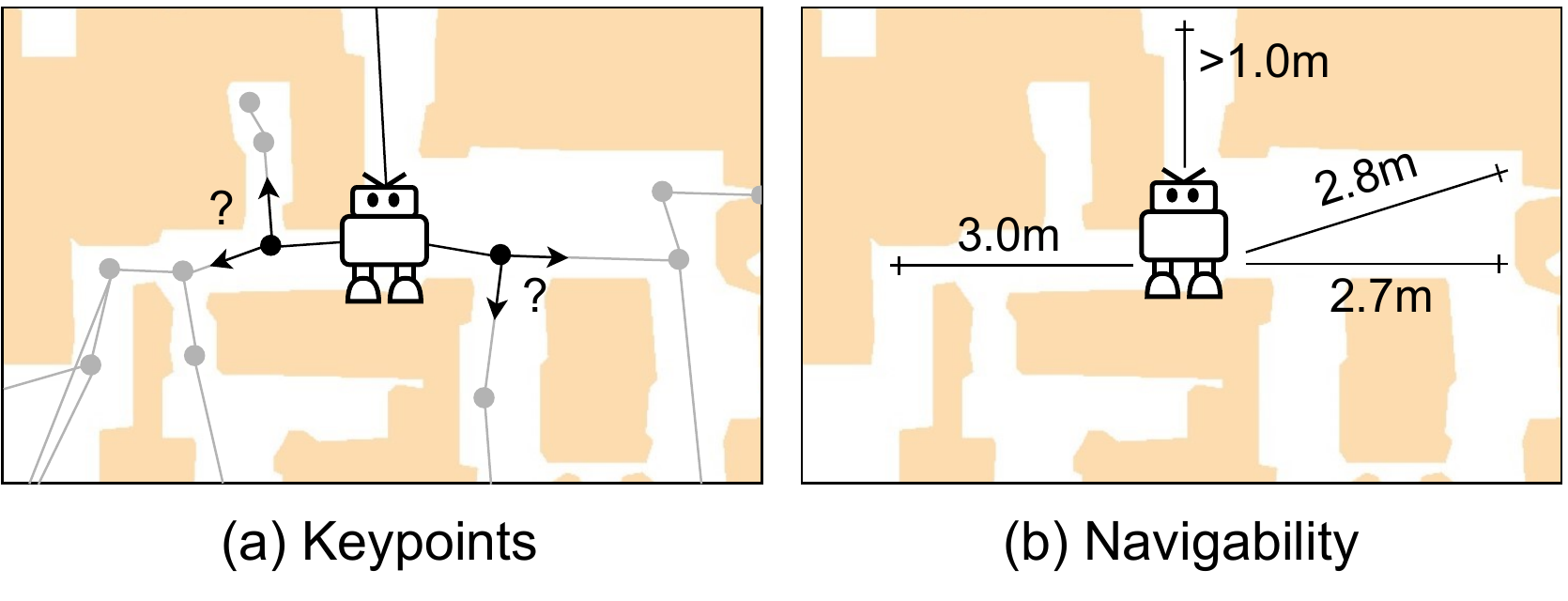}
  \vspace{-20pt}
  \caption{Keypoints versus navigability. Keypoints are closely related to the structure of the environment, which requires an agent to make a critical decision for navigation, while navigability only reflects explorable directions (and distances).}
  \label{fig:keypoint_navigability}
  \vspace{-10pt}
\end{figure}

We argue that \textit{view selection} and \textit{waypoint teleportation} are the two critical advantages resulting from high-level actions on graphs; As shown in Figure~\ref{fig:select_teleport}, view selection transfers the problem of inferring low-level controls to selecting a navigable direction, which greatly expands the agent's decision space at each time step. 
Moreover, by teleporting to a distant waypoint, the agent is able to learn and navigate very efficiently.
The connectivity graphs provide not only information about the spatial navigability, but also an exact distance to travel which transfers the agent to a location that is suitable for future decision making (see Figure~\ref{fig:keypoint_navigability}). Fried~\etal~\cite{fried2018speaker} demonstrate a huge improvement by leveraging high-level actions in MP3D, in this section, we validate the value of these two mechanisms with the RCM agent~\cite{wang2019reinforced} in Habitat-MP3D as a proof of concept.

\begin{table}[t]
    \begin{center}
    \resizebox{\columnwidth}{!}{
    \begin{tabular}{l|cc|cccc|ccccc}
        \hline \hline
        \multicolumn{1}{c|}{\multirow{2}{*}{\#}} & \multicolumn{2}{c|}{Action} & \multicolumn{4}{c|}{R2R-CE Val-Seen} & \multicolumn{5}{c}{R2R-CE Val-Unseen}\Tstrut\\
        \cline{2-12} &
        \multicolumn{1}{c}{Select} & 
        \multicolumn{1}{c|}{Teleport} & \multicolumn{1}{c}{NE$\downarrow$} &
        \multicolumn{1}{c}{nDTW$\uparrow$} & \multicolumn{1}{c}{SR$\uparrow$} & \multicolumn{1}{c|}{SPL$\uparrow$} &  \multicolumn{1}{c}{NE$\downarrow$} &
        \multicolumn{1}{c}{nDTW$\uparrow$} & \multicolumn{1}{c}{SR$\uparrow$} & \multicolumn{1}{c}{SPL$\uparrow$} & \multicolumn{1}{c}{Time}\Tstrut\\
        \hline \hline
        1  & \checkmark & \checkmark & 6.28 & 58.37 & 41.48 & 36.58 & 6.51 & 55.41 & 39.32 & 33.89 & 0.55 \\
        \hline
        2 & \checkmark &             & 5.70 & 61.06 & 46.31 & 43.20 & 6.38 & 56.27 & 37.72 & 35.02 & 1.79 \\
        3  &            & \checkmark & 8.10 & 43.19 & 23.36 & 19.85 & 7.62 & 47.69 & 28.32 & 24.64 & 0.78 \\
        \hline
        4  &  &                      & 7.20 & 50.67 & 29.53 & 27.29 & 7.54  & 49.19 & 27.29 & 24.97 & 2.61 \\         
        \hline \hline
    \end{tabular}}
    \end{center}
    \vspace{-15pt}
    \caption{Comparison of navigation in continuous environments by view selection (Select) and waypoint teleportation (Teleport) with the support of connectivity graphs.}
    \label{tab:poc_results}
    \vspace{-10pt}
\end{table}

\begin{table}[t]
    \begin{center}
    \resizebox{\columnwidth}{!}{
    \begin{tabular}{l|c|c|cccc|ccccc}
    \hline \hline
    \multicolumn{1}{c|}{\multirow{2}{*}{\#}} & \multicolumn{1}{c|}{\multirow{2}{*}{Select}} & \multicolumn{1}{c|}{\multirow{2}{*}{Dist}} & \multicolumn{4}{c|}{R2R-CE Val-Seen} & \multicolumn{5}{c}{R2R-CE Val-Unseen}\Tstrut\\
    \cline{4-12} &  &  & 
    \multicolumn{1}{c}{NE$\downarrow$} & \multicolumn{1}{c}{nDTW$\uparrow$} & \multicolumn{1}{c}{SR$\uparrow$} & \multicolumn{1}{c|}{SPL$\uparrow$} &  \multicolumn{1}{c}{NE$\downarrow$} & \multicolumn{1}{c}{nDTW$\uparrow$} &
    \multicolumn{1}{c}{SR$\uparrow$} & \multicolumn{1}{c}{SPL$\uparrow$} &  \multicolumn{1}{c}{Time}\Tstrut\\
    \hline \hline
    1 &  & 0.25 & 7.20 & 50.67 & 29.53 & 27.29 & 7.54  & 49.19 & 27.29 & 24.97 & 2.61 \\
    2 &  & 1.00 & 7.21 & 52.52 & 29.66 & 27.19 & 7.51 & 50.14 & 25.47 & 23.51 & 1.28 \\
    3 &  & 2.00 & 7.60 & 47.90 & 24.16 & 22.25 & 8.06 & 45.74 & 22.96 & 20.59 & 1.18 \\
    4 &  & 3.00 & 7.66 & 48.28 & 23.62 & 21.73 & 7.87 & 46.29 & 21.37 & 19.39 & 0.90 \\
    \hline \hline
    5 & \checkmark & 0.25 & 6.10 & 58.24 & 39.33 & 38.26 & 6.52 & 55.26 & 32.02 & 31.15 &  1.58\\
    6 & \checkmark & 1.00 & 6.88 & 53.09 & 36.11 & 33.91 & 6.85 & 52.44 & 34.81 & 32.51 & 0.46 \\
    7 & \checkmark & 2.00 & 6.56 & 52.83 & 38.52 & 35.62 & 6.96 & 50.25 & 33.05 & 30.35 & 0.27 \\
    8 & \checkmark & 3.00 & 6.75 & 51.85 & 32.21 & 29.28 & 7.00 & 49.35 & 31.00 & 28.12 & 0.20 \\
    \hline \hline
    \end{tabular}}
    \end{center}
    \vspace{-15pt}
    \caption{Navigation results using different forward distances without the connectivity graph. \textit{Dist} means the fixed forwarding distance (meters) in selected direction. \textit{Time} is the averaged inference time (seconds) per episode.}
    \label{tab:fix_waypoint_plus}
    \vspace{-15pt}
\end{table}

We first consider four different scenarios as shown in Table~\ref{tab:poc_results}: 
(1) The agent is trained and evaluated on the ground-truth connectivity graph with both view selection and waypoint teleportation.
(2) Experiment that only allows view selection; At each step, the agent's local connectivity is computed on the fly using the ground-truth graph, it can choose a navigable direction but it can only move forward 0.25 meters.
(3) As in the original VLN setup, navigable directions are not explicitly provided to the agent, but the agent is allowed to teleport forward if its heading is aligning with an edge on the connectivity graph~\cite{anderson2018vision}. We consider this setup as an experiment that allows waypoint teleportation but not view selection. 
(4) The original VLN-CE setup; The agent 
only perform low-level actions during navigation. For fairness, we apply panoramic visual input (12 cameras at 30 degrees separation) and panoramic attention to all experiments (See Appendix for more details).
Comparing M\#1\footnote{Model \#1 in the table.} to M\#3 and M\#2 to M\#4, we can see that view selection 
greatly improves the agent's performance. By knowing the navigable directions (M\#2), agent can achieve SR close to navigation on graphs (M\#1). Previous work suggests the difficulty in learning repetitive low-level actions that only return small change in view~\cite{das2018embodied}, our results further show that forwarding by teleportation greatly reduces execution time and slightly increases SR (M\#1, M\#3), while navigability is the bottleneck of learning to navigate.

Then, we investigate the influence of view selection and step size on agent's performance without considering the connectivity graph (see Table~\ref{tab:fix_waypoint_plus}). In this case, an agent navigates by choosing a view and forwarding a fixed distance. Results show that \textit{view selection} significantly boosts the agent's performance at all choices of step length (M\#5-M\#8). By adopting a larger step length, the agent also navigates much more efficiently due to less decisions to make. Note that, although an appropriate forward distance (\eg M\#6: 1 meter) leads to high performance, such value is unknown beforehand and it is unlikely to be suitable for all spatial structures. Based on the experiments, we expect to allow VLN agents in continuous spaces to take advantages of the two mechanisms, where the key is to provide the agent with candidate waypoints at any position in space.

\section{Candidate Waypoints Predictor} 
\label{sec:waypoint}

Following the above observations, in order to bridge the discrete-to-continuous gap, we propose a candidate waypoints predictor that generates virtual waypoints for agents in continuous environments. At each navigational step, the waypoints predictor infers a local sub-graph which consists of a set of edges pointing from the agent towards accessible positions in space. As a result, VLN in continuous spaces can be performed effectively using high-level actions.

\begin{figure*}[t]
  \centering
  \includegraphics[width=\textwidth]{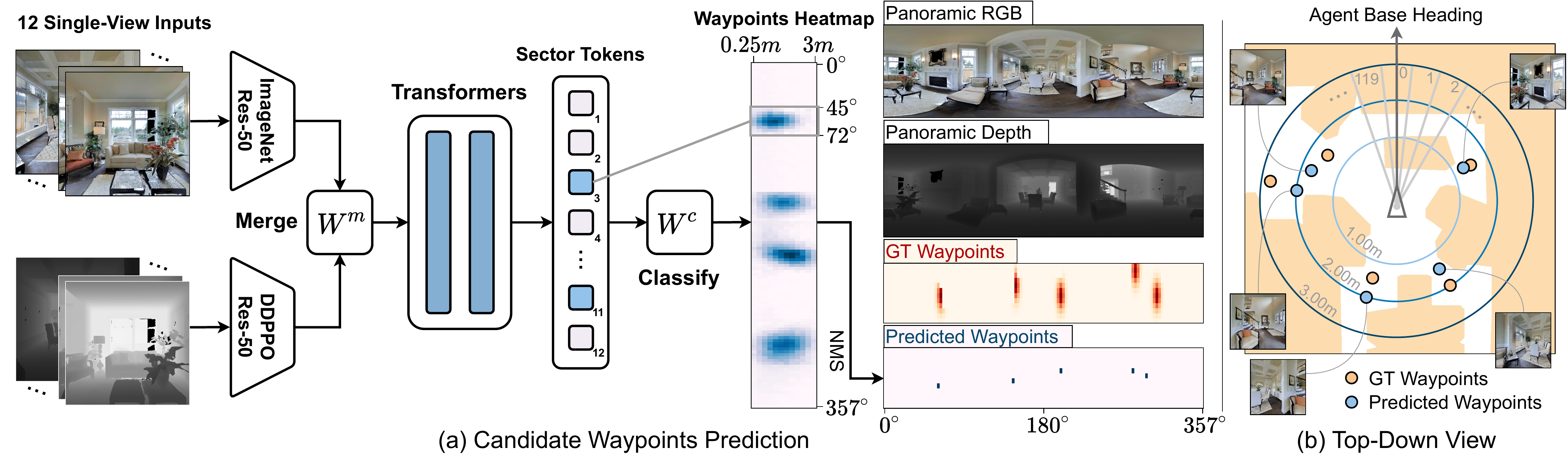}
  \caption{The candidate waypoints predictor. The module takes RGBD panoramic inputs, uses a multi-layer Transformer to model spatial relationships, and predicts the positions of waypoints in agent's neighborhood.}
  \label{fig:waypoint_prediction}
  \vspace{-10pt}
\end{figure*}

\begin{figure}[t]
\vspace{-5pt}
  \centering
  \includegraphics[width=\columnwidth]{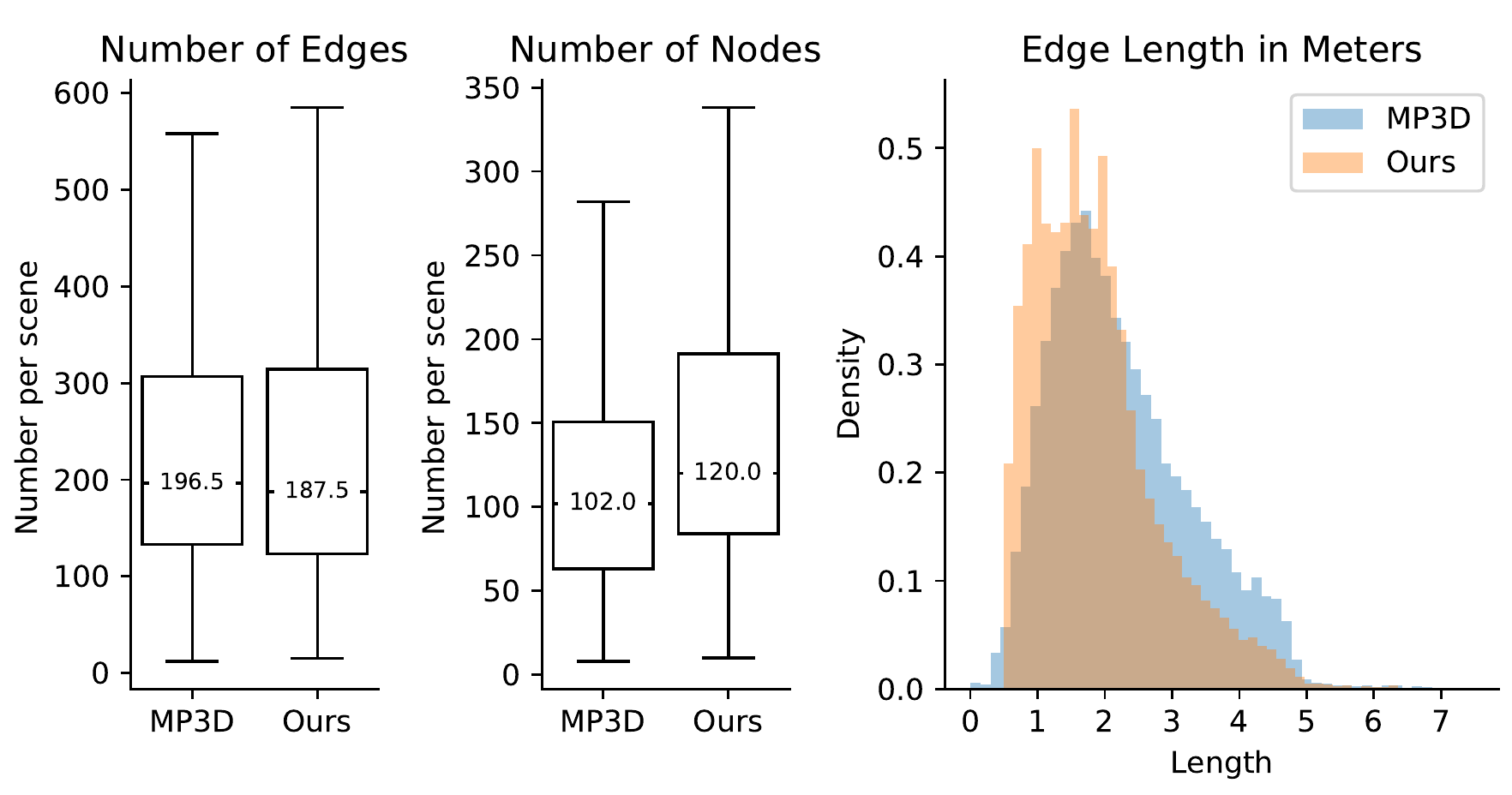}
  \vspace{-20pt}
  \caption{Statistics of connectivity graphs in MP3D and Habitat-MP3D (Ours) environments.}
  \label{fig:graph_stats}
  \vspace{-5pt}
\end{figure}

\subsection{Network Architecture and Processing}
As shown in Figure~\ref{fig:waypoint_prediction}(a), the waypoints predictor has three key modules; visual encoders, a multi-layer Transformer, and a non-linear classifier. We apply two ResNet-50~\cite{he2016deep}, one pre-trained on ImageNet~\cite{russakovsky2015imagenet} and another one pre-trained for point-goal navigation~\cite{wijmans2019dd}, for encoding RGB and depth images, respectively. The Transformer network, consisting of two layers each with 12 self-attention heads, is applied for modeling the spatial relationships between views. The classifier is a multi-layer perceptron that projects the Transformer outputs to probabilities of adjacent waypoints in space. The parameters of the visual encoders are fixed after initialization, whereas the Transformer and the classifier will be updated during training.

Given an arbitrary point in the openspace of an environment, we first gather its RGB and depth panoramas, each panorama is represented by 12 single-view images at 30 degrees separation. These images will be encoded by the visual encoders to become a sequence of RGB and depth features, denoted as $\langle{ \boldsymbol{v}^{rgb}_{1},\boldsymbol{v}^{rgb}_{2},\ldots,\boldsymbol{v}^{rgb}_{12} \mid \boldsymbol{v}^{rgb}_{i} \in {\cal V}^{rgb}}\rangle$ and $\langle{ \boldsymbol{v}^{d}_{1},\boldsymbol{v}^{d}_{2},\ldots,\boldsymbol{v}^{d}_{12} \mid \boldsymbol{v}^{d}_{i} \in {\cal V}^{d}}\rangle$, respectively. Each pair of $\boldsymbol{v}^{rgb}_{i}$ and $\boldsymbol{v}^{d}_{i}$ feature is merged by a non-linear layer $\boldsymbol{W}^{m}$, yielding $\boldsymbol{v}^{rgbd}_{i}$. All 12 visual representations $\boldsymbol{v}^{rgbd}_{i}$ are sent to the Transformer in parallel for modeling relationship and inferring adjacent waypoints. Since each single-view image is square and has 90 degrees field of view that covers three 30 degrees sectors, we constrain the self-attention of each $\boldsymbol{v}^{rgbd}_{i}$ to be performed with a single neighbor at each side, \ie with $\boldsymbol{v}^{rgbd}_{i-1}$ and $\boldsymbol{v}^{rgbd}_{i+1}$. Feature tokens outputs by the transformer $\tilde{\boldsymbol{v}}^{rgbd}_{i}$, where each contains information over a sector centered at the center of image $i$, will be fed to a classifier to predict a heatmap of 120 angles-by-12 distances. Each angle is of 3 degrees, and the distances range from 0.25 meters to 3.00 meters with 0.25 meters separation. By performing non-maximum-suppression (NMS) over the resulting heatmap, we obtain $K$ neighboring waypoints.

\subsection{Connectivity Graphs in Habitat-MP3D}

To obtain the ground-truth connectivity graph $\mathcal{G}^{*}$ for training the waypoint predictor, we adapt the connectivity graph pre-defined for MP3D ($\mathcal{G}^{MP3D}$) to fit the continuous environments in Habitat-MP3D. Notice that the two graphs only reflect partial accessibility in space, nodes defined on the graphs are more inclined to sparse keypoints in the environment which requires a navigation agent to make a crucial decision, and edges indicate directions worth exploring rather than simply navigable (Figure~\ref{fig:keypoint_navigability}). For example, it is important for an agent to predict a waypoint in front of a doorway so that the agent can decide whether to enter the room once it reaches the waypoint. In $\mathcal{G}^{MP3D}$, there exists edges that go across obstacles, which are inappropriate for predicting accessible waypoint. In contrast, all nodes and edges on $\mathcal{G}^{*}$ are defined in openspace. Additional nodes are added to ensure that the graph is connected.
Overall, the entire set of connectivity graphs contains 13,358 (10,559) nodes spread across 90 Habitat-MP3D (MP3D) environments, where each node is connected by in averaged 3.31 (4.07) edges with length in average 1.87 (2.26) meters. More statistics are shown in Figure \ref{fig:graph_stats}.

\subsection{Confirmation of Predictor Performance}
\paragraph{Dataset}
For each node on $\mathcal{G}^{*}$, we construct its local sub-graph by positioning adjacent waypoints over a discretized space. As shown in Figure~\ref{fig:waypoint_prediction}(b), we first partition the 3 meters radius range around each node into 120 3-degree sectors and 12 0.25-meter rings. Then, we assign waypoints into their corresponding partitions, and convert the discretized circle into a 120-by-12 heatmap which contains the positions of the waypoints. The heatmap is applied as prediction ground-truth, and each waypoint on the heatmap is represented as a Gaussian distribution with variance of 1.75m and 15$^{\circ}$ to allow some tolerance for the prediction. We follow the data splits in R2R to divide the connectivity graphs, using only 61 graphs (9,556 nodes) to train the predictor.

\begin{table}[t]
    \begin{center}
    \resizebox{\columnwidth}{!}{
    \begin{tabular}{l|l|cccc|cccc}
    \hline \hline
    \multicolumn{1}{c|}{\multirow{2}{*}{\#}} &
    \multicolumn{1}{c|}{\multirow{2}{*}{Model}} &  \multicolumn{4}{c|}{MP3D Train} & \multicolumn{4}{c}{MP3D Val-Unseen}\Tstrut\\
    \cline{3-10} &  &
     \multicolumn{1}{c}{$\left|\Delta\right|$} &  \multicolumn{1}{c}{\%Open $\uparrow$} &  \multicolumn{1}{c}{$d_{C}$ $\downarrow$} & \multicolumn{1}{c|}{$d_{H}$ $\downarrow$} &
     \multicolumn{1}{c}{$\left|\Delta\right|$} &  \multicolumn{1}{c}{\%Open $\uparrow$} &  \multicolumn{1}{c}{$d_{C}$ $\downarrow$} & \multicolumn{1}{c}{$d_{H}$ $\downarrow$} \Tstrut\\
    \hline \hline
    h & Baseline & 1.29 & 81.73 & 1.13 & 2.17 & 1.37 & 80.18 & 1.08 & 2.06 \\
    i & U-Net & 1.15 & 63.60 & 1.05 & 2.10 & 1.21 & 52.54 & 1.01 & 2.00 \\
    \hline
    j & Ours & 1.30 & 82.56 & 1.12 & 2.13 & 1.40 & 79.86 & 1.07 & 2.00 \\
    \hline \hline
    \end{tabular}}
    \end{center}
    \vspace{-15pt}
    \caption{Comparison of the waypoint predictors. M\#h indicates the first row (Model \#h) in the table.}
    \label{tab:waypoint_pred}
    \vspace{-15pt}
\end{table}

\paragraph{Training and Results}
We train the candidate waypoints predictor by minimizing the mean squared error between the predicted heatmaps and the ground-truth heatmaps. As shown in Table~\ref{tab:waypoint_pred}, we compare the results with (M\#h) a baseline model which does not employ a Transformer so that each single-view image is projected to a sector of waypoint scores independently, and (M\#i) a convolutional U-Net~\cite{ronneberger2015u} which is similar to the sub-goal module in Sim2Real~\cite{anderson2020sim}. We set the maximum number of prediction from each heatmap to be 5. All models are trained on a NVIDIA 3090 GPU with a learning rate of 10$^{-6}$ and batch size 64 using the AdamW optimizer~\cite{loshchilov2017decoupled}.

We evaluate the performance of the waypoint predictors using four metrics: $\left|\Delta\right|$ measures the difference in number of target waypoints and predicted waypoints. \%Open measures the ratio of predicted waypoints that is in open space (not hindered by any obstacle). $d_{C}$ and $d_{H}$ are the Chamfer distance and the Hausdorff distance, respectively, which are commonly used metrics for measuring the distance between point clouds. As shown in Table~\ref{tab:waypoint_pred}, U-Net results in the waypoints closest to the ground-truths, but a large portion of the predictions are block by obstacles, which is inappropriate for navigation. Comparing to the baseline, our model with the Transformer achieves similar \%Open but lower distances, which will be applied in the following sections.

\begin{table*}[!htp]
  \begin{center}
  \resizebox{1.00\textwidth}{!}{
  \begin{tabular}{l|c|rr|rr|rrcrr|rrcrc}
    \hline \hline
    \multicolumn{1}{c|}{\multirow{2}{*}{Methods}} & \multicolumn{1}{c|}{\multirow{2}{*}{\#}} & \multicolumn{2}{c|}{Train Connectivity} & \multicolumn{2}{c|}{Val Connectivity} & \multicolumn{5}{c|}{R2R-CE Val-Seen} &
    \multicolumn{5}{c}{R2R-CE Val-Unseen} \\
    \cline{3-16} & & 
    \multicolumn{1}{c}{Graph} & \multicolumn{1}{c|}{Predictor} & \multicolumn{1}{c}{Graph} & \multicolumn{1}{c|}{Predictor} & \multicolumn{1}{c}{TL} & \multicolumn{1}{c}{NE$\downarrow$} & \multicolumn{1}{c}{nDTW$\uparrow$} & \multicolumn{1}{c}{SR$\uparrow$} & \multicolumn{1}{c|}{SPL$\uparrow$} & \multicolumn{1}{c}{TL} & \multicolumn{1}{c}{NE$\downarrow$} & \multicolumn{1}{c}{nDTW$\uparrow$} & \multicolumn{1}{c}{SR$\uparrow$} & \multicolumn{1}{c}{SPL$\uparrow$}\Tstrut\\
    \hline \hline
    \multicolumn{1}{c|}{\multirow{5}{*}{CMA}} & 1 & \multicolumn{1}{c|}{\multirow{1}{*}{\checkmark}} & \multicolumn{1}{c|}{\multirow{1}{*}{}} & \multicolumn{1}{c|}{\multirow{1}{*}{\checkmark}} & \multicolumn{1}{c|}{\multirow{1}{*}{}} & 10.21 & 6.28 & 58.37 & 41.48 & \multicolumn{1}{c|}{\multirow{1}{*}{36.58}} & 10.47 & 6.51 & 55.41 & 39.32 & 33.89 \\
     & 2 & \multicolumn{1}{c|}{\multirow{1}{*}{\checkmark}} & \multicolumn{1}{c|}{\multirow{1}{*}{}} & \multicolumn{1}{c|}{\multirow{1}{*}{}} & \multicolumn{1}{c|}{\multirow{1}{*}{Freeze}} & 8.71 & 6.83 & 53.16 & 33.83 & \multicolumn{1}{c|}{\multirow{1}{*}{30.23}} & 8.28 & 6.81 & 52.86 & 33.50 & 29.92 \\
     \cline{3-16}
     & 3 & \multicolumn{1}{c|}{\multirow{1}{*}{}} & \multicolumn{1}{c|}{\multirow{1}{*}{Freeze}} & \multicolumn{1}{c|}{\multirow{1}{*}{}} & \multicolumn{1}{c|}{\multirow{1}{*}{Freeze}} & 9.39 & 5.81 & 59.06 & 43.89 & \multicolumn{1}{c|}{\multirow{1}{*}{40.20}} & 8.77 & 6.50 & 54.34 & 37.49 & 33.90 \\
     & 4 & \multicolumn{1}{c|}{\multirow{1}{*}{}} & \multicolumn{1}{c|}{\multirow{1}{*}{Augmented}} & \multicolumn{1}{c|}{\multirow{1}{*}{}} & \multicolumn{1}{c|}{\multirow{1}{*}{Freeze}} & \cellcolor{lightgray}9.57 & \cellcolor{lightgray}5.36 & \cellcolor{lightgray}61.46 & \cellcolor{lightgray}48.05 & \multicolumn{1}{c|}{\multirow{1}{*}{\cellcolor{lightgray}43.89}} &  \cellcolor{lightgray}9.11 & \cellcolor{lightgray}6.19 & \cellcolor{lightgray}55.56 & \cellcolor{lightgray}40.80 & \cellcolor{lightgray}36.73 \\
     \cline{3-16}
     & 5 & \multicolumn{4}{c|}{\multirow{1}{*}{Low-Level Actions}} & 8.87 & 7.20 & 50.67 & 29.53 & \multicolumn{1}{c|}{\multirow{1}{*}{27.29}} & 8.22 & 7.54 & 49.19 & 27.29 & 24.97 \\
    \hline \hline
    \multicolumn{1}{c|}{\multirow{5}{*}{\vlnbert}} & 6 & \multicolumn{1}{c|}{\multirow{1}{*}{\checkmark}} & \multicolumn{1}{c|}{\multirow{1}{*}{}} & \multicolumn{1}{c|}{\multirow{1}{*}{\checkmark}} & \multicolumn{1}{c|}{\multirow{1}{*}{}} & 14.21 & 4.63 & 61.46 & 52.21 & \multicolumn{1}{c|}{\multirow{1}{*}{42.53}} & 14.34 & 5.22 & 57.71 & 48.89 & 40.36 \\
     & 7 & \multicolumn{1}{c|}{\multirow{1}{*}{\checkmark}} & \multicolumn{1}{c|}{\multirow{1}{*}{}} & \multicolumn{1}{c|}{\multirow{1}{*}{}} & \multicolumn{1}{c|}{\multirow{1}{*}{Freeze}} & 11.42 & 5.62 & 56.19 & 43.22 & \multicolumn{1}{c|}{\multirow{1}{*}{37.10}} & 11.22 & 5.91 & 53.35 & 39.94 & 34.42 \\
     \cline{3-16}
     & 8 & \multicolumn{1}{c|}{\multirow{1}{*}{}} & \multicolumn{1}{c|}{\multirow{1}{*}{Freeze}} & \multicolumn{1}{c|}{\multirow{1}{*}{}} & \multicolumn{1}{c|}{\multirow{1}{*}{Freeze}} & 11.24 & 5.06 & 59.46 & 49.40 & \multicolumn{1}{c|}{\multirow{1}{*}{43.43}} & 12.12 & 5.68 & 53.50 & 42.96 & 38.53 \\
     & 9 & \multicolumn{1}{c|}{\multirow{1}{*}{}} & \multicolumn{1}{c|}{\multirow{1}{*}{Augmented}} & \multicolumn{1}{c|}{\multirow{1}{*}{}} & \multicolumn{1}{c|}{\multirow{1}{*}{Freeze}} & \cellcolor{lightgray}11.83 & \cellcolor{lightgray}5.07 & \cellcolor{lightgray}59.18 &  \cellcolor{lightgray}52.35 & \multicolumn{1}{c|}{\multirow{1}{*}{\cellcolor{lightgray}46.08}} & \cellcolor{lightgray}11.85 & \cellcolor{lightgray}5.52 & \cellcolor{lightgray}54.20 & \cellcolor{lightgray}45.19 & \cellcolor{lightgray}39.91 \\
     \cline{3-16}
     & 10 & \multicolumn{4}{c|}{\multirow{1}{*}{Low-Level Actions}} & 8.47 & 7.40 & 48.07 & 26.85 & \multicolumn{1}{c|}{\multirow{1}{*}{25.19}} & 7.42 & 7.66 & 47.71 & 23.19 & 21.67 \\
    \hline \hline
  \end{tabular}}
\end{center}
\vspace{-10pt}
\caption{Agent performance in R2R-CE. 4\% of samples which does not have a valid discrete path are deleted (details in Appendix), and all ground-truth paths are set to the shortest paths on graph for a fair comparison between navigation with and without the connectivity graph. \textit{Graph} and \textit{Predictor} means applying ground-truth graph and using either \textit{Freezed} or \textit{Augmented} predicted waypoints, respectively. 
}
\vspace{-8pt}
\label{tab:main_results_wp}
\end{table*}

\section{Bridging the Discrete to Continuous Gap} 
\label{sec:adaption}

Thanks to the waypoints predictor, agents designed for high-level actions and pre-trained on graphs~\cite{hong2020recurrent} can be transferred to continuous environments without any pre-defined graph. In this section, we demonstrate that the performance gap between the two navigation setups is significantly reduced. Furthermore, we show that using augmented waypoints for training leads to more robust agents, and new state-of-the-art results can be achieved on the benchmarking R2R-CE~\cite{anderson2018vision,krantz2020navgraph} and RxR-CE~\cite{anderson2020rxr} datasets.

\subsection{Setups}

\paragraph{Datasets}
We evaluate the performance of agents on two datasets, R2R-CE~\cite{anderson2018vision,krantz2020navgraph} and RxR-CE~\cite{anderson2020rxr}, using the Habitat simulator~\cite{savva2019habitat}. Both datasets are collected based on the discrete MP3D environments, including 61 environments for training, 11 for unseen validation and 18 reserved environments for testing. SPL~\cite{anderson2018evaluation} and nDTW~\cite{ilharco2019general} are the main evaluation metrics for R2R-CE and RxR-CE, respectively.

\paragraph{Experiments}
Our main experiment consists of two parts; First, we compare the agent's performance with and without using the connectivity graphs $\mathcal{G}^{*}$ (Table~\ref{tab:main_results_wp}). For fairness, we filter out 4\% of samples from the original R2R-CE which has a continuous ground-truth path that cannot be converted to a valid discrete path on $\mathcal{G}^{*}$ that honors the navigation instruction, and the start and end points of some paths have been slightly adjusted so that all point are on $\mathcal{G}^{*}$. Second, we compare our method to previous approaches (Table~\ref{tab:main_results_r2r} and Table~\ref{tab:main_results_rxr}). In this case, we evaluate our agents on the untouched validation and test data of R2R-CE and RxR-CE.

\paragraph{Navigators}
We experiment with two navigators that have been widely applied in previous works while distinct in network architecture or navigation method. (1) CMA~\cite{wang2019reinforced} is a simple sequence-to-sequence network with visual and language attentions as explained in Eq.\ref{eqn:state_lstm} and Eq.\ref{eqn:rcm_discrete_action}. (2) The Recurrent VLN-BERT (\vlnbert)~\cite{hong2020recurrent} is a Transformer-based network pre-trained for discrete VLN on R2R, it leverages the pre-defined \texttt{[CLS]} token in BERT as the agent's state and applies it for multi-modal attentions. For CMA, the entire network, including the language embeddings, is trained from scratch. Whereas the \vlnbert~is initialized from the pre-trained Transformers~\cite{hao2020towards}. We refer the readers to Appendix for details about the network architectures and the different configurations of agents in R2R-CE and RxR-CE.

\begin{table*}[ht!]
  \begin{center}
  \resizebox{1.00\textwidth}{!}{
  \begin{tabular}{l|cccccc|cccccc|ccccc}
    \hline \hline
    \multicolumn{1}{c|}{\multirow{2}{*}{Methods}} &  \multicolumn{6}{c|}{R2R-CE Val-Seen} &
    \multicolumn{6}{c|}{R2R-CE Val-Unseen} &
    \multicolumn{5}{c}{R2R-CE Test-Unseen} \\
    \cline{2-18} & 
    \multicolumn{1}{c}{TL} & \multicolumn{1}{c}{NE$\downarrow$} &
    \multicolumn{1}{c}{nDTW$\uparrow$} &
    \multicolumn{1}{c}{OSR$\uparrow$} & \multicolumn{1}{c}{SR$\uparrow$} & \multicolumn{1}{c|}{SPL$\uparrow$} &  \multicolumn{1}{c}{TL} & \multicolumn{1}{c}{NE$\downarrow$} &
    \multicolumn{1}{c}{nDTW$\uparrow$} &
    \multicolumn{1}{c}{OSR$\uparrow$} & \multicolumn{1}{c}{SR$\uparrow$} & \multicolumn{1}{c|}{SPL$\uparrow$} &  \multicolumn{1}{c}{TL} & \multicolumn{1}{c}{NE$\downarrow$} &
    \multicolumn{1}{c}{OSR$\uparrow$} & \multicolumn{1}{c}{SR$\uparrow$} & \multicolumn{1}{c}{SPL$\uparrow$} \Tstrut\\
    \hline \hline
    VLN-CE~\cite{krantz2020navgraph} & 9.26 & 7.12 & 54 & 46 & 37 & 35 & 8.64 & 7.37 & 51 & 40 & 32 & 30 & 8.85 & 7.91 & 36 & 28 & 25 \\
    LAW~\cite{raychaudhuri2021law} & -- & -- & 58 & -- & 40 & 37 & -- & -- & 54 & -- & 35 & 31 & -- & -- & -- & -- & -- \\
    SASRA~\cite{irshad2021sasra} & 8.89 & 7.17 & 53 & -- & 36 & 34 & 7.89 & 8.32 & 47 & -- & 24 & 22 & -- & -- & -- & -- & -- \\
    Waypoint Models~\cite{krantz2021waypoint} & 8.54 & 5.48 & -- & 53 & 46 & 43 & 7.62 & 6.31 & -- & 40 & 36 & 34 & 8.02 & 6.65 & 37 & 32 & 30 \\
    \hline
    Ours (CMA) & 11.47 & 5.20 & 61 & 61 & 51 & 45 & 10.90 & 6.20 & 55 & 52 & 41 & 36 & 11.85 & \textbf{6.30} & \textbf{49} & \textbf{38} & \textbf{33} \\
    Ours (\vlnbert) & 12.50 & 5.02 & 58 & 59 & 50 & 44 & 12.23 & 5.74 & 54 & 53 & 44 & 39 & 13.31 & \high{5.89} & \high{51} & \high{42} & \high{36} \\
    \hline \hline
  \end{tabular}}
\end{center}
\vspace{-10pt}
\caption{Comparison on agent performance in R2R-CE. Samples in all data splits are identical as in VLN-CE~\cite{krantz2020navgraph}.}
\vspace{-10pt}
\label{tab:main_results_r2r}
\end{table*}

\paragraph{Training}
All experiments are conducted on a single NVIDIA RTX 3090 GPU based on the PyTorch framework~\cite{paszke2019pytorch} and the Habitat simulator~\cite{savva2019habitat}. We apply a simple cross-entropy loss between the ground-truth actions and agent's predictions as the learning objective, which is minimized with an AdamW optimizer~\cite{loshchilov2017decoupled} during training. All the agents are trained using schedule sampling~\cite{bengio2015scheduled} with a decay frequency (moving from teacher forcing to student forcing) of per 5 epochs. As in previous work~\cite{krantz2020navgraph}, for the CMA agents, the decay ratio is set to 0.75 and the learning rate is $10^{-4}$. For the \vlnbert, the decay ratio is set to 0.50 (0.75 in RxR-CE due to the longer paths) and the learning rate is $10^{-5}$. For agents which navigate on graphs, we consider the candidate node which has the shortest Dijkstra path to the target as ground-truth. For agents which navigate with predicted waypoints, we use a trained candidate waypoints predictor (from \S\ref{sec:waypoint}) to generate accessible sub-goals to allow the agents perform \textit{view selection}. The ground-truth waypoint is the one which has the shortest geodesic distance to the target and to the sub-goal in R2R-CE and RxR-CE, respectively (see Appendix). Once a waypoint is chosen, instead of teleporting directly to the waypoint, we decompose such high-level goal to low-level controls, then, ask the agent to execute those controls progressively.

\subsection{Main Results}

\paragraph{Bridge the Gap}
As shown in Table~\ref{tab:main_results_wp}, agents navigate on the ground-truth graphs (M\#1, M\#6) perform significantly better than agents navigate with low-level actions (M\#5, M\#10), scoring 8.92\% and 18.69\% higher SPL scores for the CMA and \vlnbert, respectively.
M\#2 and M\#7 show that when the connectivity graph is absent, using a waypoints predictor can bring up the agents' performance by a large margin. Moreover, if the agents are trained on the predicted waypoints (M\#3, M\#8), so that the navigators can learn to adapt the patterns of generated waypoints, then agents can achieve similar performance as navigating on the ground-truth graphs -- For CMA, the agent scores the same SPL as M\#1. For \vlnbert, the gaps of SR and SPL are reduced by 77\% and 90\%, respectively. 
Trace back to early experiments with fixed step length, there are about 15\% of steps executed by M\#6 in Table~\ref{tab:fix_waypoint_plus} collides with obstacles, while this number is only 7\% once the waypoints predictor is introduced, which again reflects the importance of the predictor.
However, in some rare cases our waypoints predictor may miss a direction connecting the agent to the destination (such as traversing stairs) resulting in a navigation failure.
Sim2Real-VLN~\cite{anderson2020sim} suggests that the quality of the predicted waypoint candidates is a critical bottleneck of the agent's performance, whereas our results indicate that overall our waypoints predictor is generalizable to new environments, which successfully allows the agents to learn and act with high-level actions. We refer to the Appendix for visualization of the predicted waypoints and navigation trajectories, as well as further discussion on limitations.\vspace{-10pt}

\begin{figure}[t]
  \centering
  \includegraphics[width=\columnwidth]{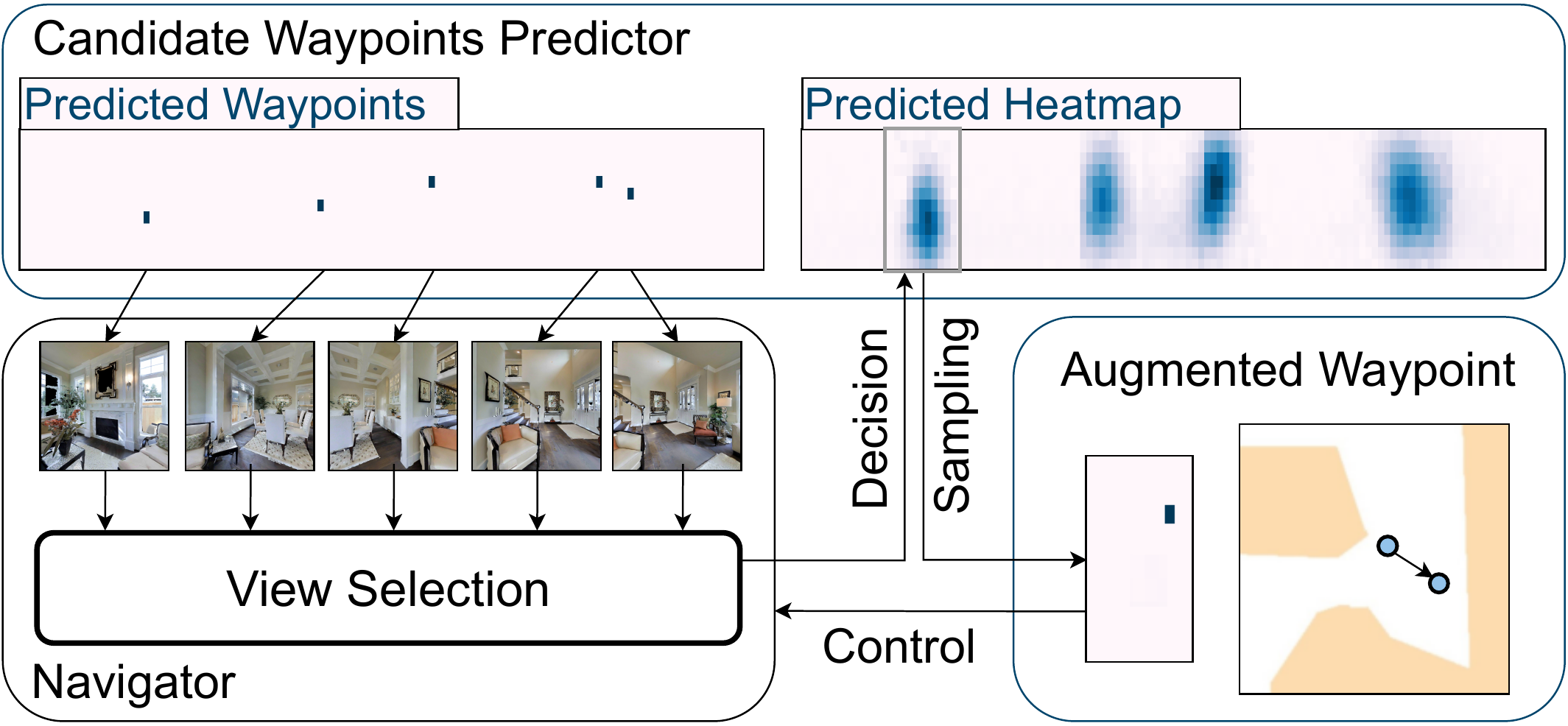}
  \vspace{-15pt}
  \caption{Waypoint Augmentation.}
  \label{fig:waypoint_aug}
  \vspace{-15pt}
\end{figure}

\paragraph{Waypoint Augmentation}
We propose to augment the waypoints produced by the candidate waypoints predictor while training the agent to enhance the agent's generalization ability. As shown in Figure~\ref{fig:waypoint_aug}, the method samples a new waypoint from a patch of heatmap which corresponds to the agent's selected view. The sampled waypoints will take the agent to new positions in space, which requires the agent to observe diverse views, travel with different step lengths and interacts with different obstacles to complete the same navigation task. Results in Table~\ref{tab:main_results_wp} show that the agents' performances are further improved by training with augmented waypoints (M\#4, M\#9); the SPL of CMA in unseen environments even exceeds CMA navigating on graphs (M\#1). These results indicate the effectiveness of augmenting waypoints when applying high-level actions.

\begin{table}[ht!]
  \begin{center}
  \resizebox{0.95\columnwidth}{!}{
  \begin{tabular}{l|rrrrrr}
    \hline \hline
    \multicolumn{1}{c|}{\multirow{2}{*}{Methods}} &  
    \multicolumn{6}{c}{RxR-CE Test-Unseen} \\
    \cline{2-7} & 
    \multicolumn{1}{c}{TL} & \multicolumn{1}{c}{NE$\downarrow$} &
    \multicolumn{1}{c}{SR$\uparrow$} & \multicolumn{1}{c}{SPL$\uparrow$} &
    \multicolumn{1}{c}{nDTW$\uparrow$} &
    \multicolumn{1}{c}{SDTW$\uparrow$} \Tstrut\\
    \hline \hline
    VLN-CE~\cite{krantz2020navgraph} & 7.33 & 12.1 & 13.93 & 11.96 & 30.86 & 11.01 \\
    \hline
    Ours (CMA) & 20.04 & \high{10.4} & \textbf{24.08} & \textbf{19.07} & \high{37.39} & \textbf{18.65} \\
    Ours (\vlnbert) & 20.09 & \high{10.4} & \high{24.85} & \high{19.61} & \textbf{37.30} & \high{19.05} \\
    \hline \hline
  \end{tabular}}
\end{center}
\vspace{-12pt}
\caption{Comparison on agent performance in RxR-CE.}
\vspace{-15pt}
\label{tab:main_results_rxr}
\end{table}

\paragraph{Comparison to SoTA}
We train the agents with predicted waypoints on the original R2R-CE and RxR-CE datasets, respectively, and compare the results with the previous state-of-the-arts\footnote{R2R-CE Leaderboard: \href{https://eval.ai/web/challenges/challenge-page/719/leaderboard/1966}{https://eval.ai/web/challenges/challenge-page/719/leaderboard/1966}, RxR-CE Leaderboard:\\ \href{https://ai.google.com/research/rxr/habitat}{https://ai.google.com/research/rxr/habitat} \label{footnote_leaderboard}} (Table~\ref{tab:main_results_r2r}, Table~\ref{tab:main_results_rxr}). Our method significantly outperforms previous methods across all dataset splits. On the test server of R2R-CE, our CMA agent improves over the CMA in Waypoint Models~\cite{krantz2021waypoint} by 6\% SR and 3\% SPL. It is also worth mentioning that comparing to the Waypoint Models which applies DDPPO~\cite{wijmans2019dd} and 64 GPUs (5 days) for training the agent, our approach enables an efficient imitation learning, which drastically reduces the training cost to a single GPU (3.5 days) while achieving better results. On the multilingual RxR-CE dataset, both CMA and \vlnbert~achieve more than 10\% higher SR and 6\% higher nDTW than the previous best model. 

\section{Conclusion} 
\label{sec:conclusion}

In this paper, we introduce a candidate waypoints predictor to produce accessible waypoints in space, which enables agents designed for discrete environments to learn and navigate in continuous environments with high-level actions. Experiments show that our method is generalizable for different agents and to unseen environments, it greatly bridges the discrete-to-continuous gap and achieves new state-of-the-art performances on the R2R-CE and the RxR-CE datasets. We believe this work is an important step towards taking VLN research in MP3D to the more realistic scenarios, including simulation in continuous environments and even robots in the real-world. For future work, weakly supervised and language-conditioned waypoint prediction could be explored. Moreover, the idea of predicting adjacent waypoints and performing \textit{navigable-view selection} have the great potential to be applied in a wide range of research such as PointNav~\cite{savva2019habitat}, ObjectNav~\cite{batra2020objectnav}, Audio-Visual Nav~\cite{chen2020soundspaces} and embodied task-completion problems~\cite{shridhar2020alfred}.

{\small
\bibliographystyle{ieee_fullname}
\bibliography{egbib}

\begin{thebibliography}{10}\itemsep=-1pt

\bibitem{anderson2018evaluation}
Peter Anderson, Angel Chang, Devendra~Singh Chaplot, Alexey Dosovitskiy,
  Saurabh Gupta, Vladlen Koltun, Jana Kosecka, Jitendra Malik, Roozbeh
  Mottaghi, Manolis Savva, et~al.
\newblock On evaluation of embodied navigation agents.
\newblock {\em arXiv preprint arXiv:1807.06757}, 2018.

\bibitem{anderson2020sim}
Peter Anderson, Ayush Shrivastava, Joanne Truong, Arjun Majumdar, Devi Parikh,
  Dhruv Batra, and Stefan Lee.
\newblock Sim-to-real transfer for vision-and-language navigation.
\newblock In {\em Conference on Robot Learning}, pages 671--681. PMLR, 2021.

\bibitem{anderson2018vision}
Peter Anderson, Qi Wu, Damien Teney, Jake Bruce, Mark Johnson, Niko
  S{\"u}nderhauf, Ian Reid, Stephen Gould, and Anton van~den Hengel.
\newblock Vision-and-language navigation: Interpreting visually-grounded
  navigation instructions in real environments.
\newblock In {\em Proceedings of the IEEE Conference on Computer Vision and
  Pattern Recognition}, pages 3674--3683, 2018.

\bibitem{batra2020objectnav}
Dhruv Batra, Aaron Gokaslan, Aniruddha Kembhavi, Oleksandr Maksymets, Roozbeh
  Mottaghi, Manolis Savva, Alexander Toshev, and Erik Wijmans.
\newblock Objectnav revisited: On evaluation of embodied agents navigating to
  objects.
\newblock {\em arXiv preprint arXiv:2006.13171}, 2020.

\bibitem{bengio2015scheduled}
Samy Bengio, Oriol Vinyals, Navdeep Jaitly, and Noam Shazeer.
\newblock Scheduled sampling for sequence prediction with recurrent neural
  networks.
\newblock In {\em Proceedings of the 28th International Conference on Neural
  Information Processing Systems-Volume 1}, pages 1171--1179, 2015.

\bibitem{chang2017matterport3d}
Angel Chang, Angela Dai, Thomas Funkhouser, Maciej Halber, Matthias Niebner,
  Manolis Savva, Shuran Song, Andy Zeng, and Yinda Zhang.
\newblock Matterport3d: Learning from rgb-d data in indoor environments.
\newblock In {\em 2017 International Conference on 3D Vision (3DV)}, pages
  667--676. IEEE, 2017.

\bibitem{chaplot2020learning}
Devendra~Singh Chaplot, Dhiraj Gandhi, Saurabh Gupta, Abhinav Gupta, and Ruslan
  Salakhutdinov.
\newblock Learning to explore using active neural slam.
\newblock In {\em International Conference on Learning Representations}, 2019.

\bibitem{chaplot2020neural}
Devendra~Singh Chaplot, Ruslan Salakhutdinov, Abhinav Gupta, and Saurabh Gupta.
\newblock Neural topological slam for visual navigation.
\newblock In {\em Proceedings of the IEEE/CVF Conference on Computer Vision and
  Pattern Recognition}, pages 12875--12884, 2020.

\bibitem{chen2020soundspaces}
Changan Chen, Unnat Jain, Carl Schissler, Sebastia Vicenc~Amengual Gari, Ziad
  Al-Halah, Vamsi~Krishna Ithapu, Philip Robinson, and Kristen Grauman.
\newblock Soundspaces: Audio-visual navigation in 3d environments.
\newblock In {\em Computer Vision--ECCV 2020: 16th European Conference,
  Glasgow, UK, August 23--28, 2020, Proceedings, Part VI 16}, pages 17--36.
  Springer, 2020.

\bibitem{chen2020learning}
Changan Chen, Sagnik Majumder, Ziad Al-Halah, Ruohan Gao, Santhosh~Kumar
  Ramakrishnan, and Kristen Grauman.
\newblock Learning to set waypoints for audio-visual navigation.
\newblock In {\em International Conference on Learning Representations}, 2020.

\bibitem{chen2019touchdown}
Howard Chen, Alane Suhr, Dipendra Misra, Noah Snavely, and Yoav Artzi.
\newblock Touchdown: Natural language navigation and spatial reasoning in
  visual street environments.
\newblock In {\em Proceedings of the IEEE Conference on Computer Vision and
  Pattern Recognition}, pages 12538--12547, 2019.

\bibitem{chen2021topological}
Kevin Chen, Junshen~K Chen, Jo Chuang, Marynel V{\'a}zquez, and Silvio
  Savarese.
\newblock Topological planning with transformers for vision-and-language
  navigation.
\newblock In {\em Proceedings of the IEEE/CVF Conference on Computer Vision and
  Pattern Recognition}, pages 11276--11286, 2021.

\bibitem{chen2019behavioral}
Kevin Chen, Juan~Pablo de Vicente, Gabriel Sepulveda, Fei Xia, Alvaro Soto,
  Marynel V{\'a}zquez, and Silvio Savarese.
\newblock A behavioral approach to visual navigation with graph localization
  networks.
\newblock {\em arXiv preprint arXiv:1903.00445}, 2019.

\bibitem{chen2018learning}
Tao Chen, Saurabh Gupta, and Abhinav Gupta.
\newblock Learning exploration policies for navigation.
\newblock In {\em International Conference on Learning Representations}, 2018.

\bibitem{das2018embodied}
Abhishek Das, Samyak Datta, Georgia Gkioxari, Stefan Lee, Devi Parikh, and
  Dhruv Batra.
\newblock Embodied question answering.
\newblock In {\em Proceedings of the IEEE Conference on Computer Vision and
  Pattern Recognition}, pages 1--10, 2018.

\bibitem{deng2020evolving}
Zhiwei Deng, Karthik Narasimhan, and Olga Russakovsky.
\newblock Evolving graphical planner: Contextual global planning for
  vision-and-language navigation.
\newblock {\em arXiv preprint arXiv:2007.05655}, 2020.

\bibitem{devlin2019bert}
Jacob Devlin, Ming-Wei Chang, Kenton Lee, and Kristina Toutanova.
\newblock Bert: Pre-training of deep bidirectional transformers for language
  understanding.
\newblock In {\em Proceedings of the 2019 Conference of the North American
  Chapter of the Association for Computational Linguistics: Human Language
  Technologies, Volume 1 (Long and Short Papers)}, pages 4171--4186, 2019.

\bibitem{fried2018speaker}
Daniel Fried, Ronghang Hu, Volkan Cirik, Anna Rohrbach, Jacob Andreas,
  Louis-Philippe Morency, Taylor Berg-Kirkpatrick, Kate Saenko, Dan Klein, and
  Trevor Darrell.
\newblock Speaker-follower models for vision-and-language navigation.
\newblock In {\em Advances in Neural Information Processing Systems}, pages
  3314--3325, 2018.

\bibitem{fu2020counterfactual}
Tsu-Jui Fu, Xin~Eric Wang, Matthew~F Peterson, Scott~T Grafton, Miguel~P
  Eckstein, and William~Yang Wang.
\newblock Counterfactual vision-and-language navigation via adversarial path
  sampler.
\newblock In {\em European Conference on Computer Vision}, pages 71--86.
  Springer, 2020.

\bibitem{gupta2017cognitive}
Saurabh Gupta, James Davidson, Sergey Levine, Rahul Sukthankar, and Jitendra
  Malik.
\newblock Cognitive mapping and planning for visual navigation.
\newblock In {\em Proceedings of the IEEE Conference on Computer Vision and
  Pattern Recognition}, pages 2616--2625, 2017.

\bibitem{hao2020towards}
Weituo Hao, Chunyuan Li, Xiujun Li, Lawrence Carin, and Jianfeng Gao.
\newblock Towards learning a generic agent for vision-and-language navigation
  via pre-training.
\newblock In {\em Proceedings of the IEEE/CVF Conference on Computer Vision and
  Pattern Recognition}, pages 13137--13146, 2020.

\bibitem{he2016deep}
Kaiming He, Xiangyu Zhang, Shaoqing Ren, and Jian Sun.
\newblock Deep residual learning for image recognition.
\newblock In {\em Proceedings of the IEEE conference on computer vision and
  pattern recognition}, pages 770--778, 2016.

\bibitem{hochreiter1997long}
Sepp Hochreiter and J{\"u}rgen Schmidhuber.
\newblock Long short-term memory.
\newblock {\em Neural computation}, 9(8):1735--1780, 1997.

\bibitem{hong2020graph}
Yicong Hong, Cristian Rodriguez, Yuankai Qi, Qi Wu, and Stephen Gould.
\newblock Language and visual entity relationship graph for agent navigation.
\newblock {\em Advances in Neural Information Processing Systems}, 33, 2020.

\bibitem{hong2020recurrent}
Yicong Hong, Qi Wu, Yuankai Qi, Cristian Rodriguez-Opazo, and Stephen Gould.
\newblock A recurrent vision-and-language bert for navigation.
\newblock In {\em Proceedings of the IEEE/CVF Conference on Computer Vision and
  Pattern Recognition (CVPR)}, pages 1643--1653, June 2021.

\bibitem{hu2019you}
Ronghang Hu, Daniel Fried, Anna Rohrbach, Dan Klein, Trevor Darrell, and Kate
  Saenko.
\newblock Are you looking? grounding to multiple modalities in
  vision-and-language navigation.
\newblock In {\em Proceedings of the 57th Annual Meeting of the Association for
  Computational Linguistics}, pages 6551--6557, 2019.

\bibitem{ilharco2019general}
Gabriel Ilharco, Vihan Jain, Alexander Ku, Eugene Ie, and Jason Baldridge.
\newblock General evaluation for instruction conditioned navigation using
  dynamic time warping.
\newblock {\em arXiv preprint arXiv:1907.05446}, 2019.

\bibitem{irshad2021hierarchical}
Muhammad~Zubair Irshad, Chih-Yao Ma, and Zsolt Kira.
\newblock Hierarchical cross-modal agent for robotics vision-and-language
  navigation.
\newblock {\em arXiv preprint arXiv:2104.10674}, 2021.

\bibitem{irshad2021sasra}
Muhammad~Zubair Irshad, Niluthpol~Chowdhury Mithun, Zachary Seymour, Han-Pang
  Chiu, Supun Samarasekera, and Rakesh Kumar.
\newblock Sasra: Semantically-aware spatio-temporal reasoning agent for
  vision-and-language navigation in continuous environments.
\newblock {\em arXiv preprint arXiv:2108.11945}, 2021.

\bibitem{ke2019tactical}
Liyiming Ke, Xiujun Li, Yonatan Bisk, Ari Holtzman, Zhe Gan, Jingjing Liu,
  Jianfeng Gao, Yejin Choi, and Siddhartha Srinivasa.
\newblock Tactical rewind: Self-correction via backtracking in
  vision-and-language navigation.
\newblock In {\em Proceedings of the IEEE Conference on Computer Vision and
  Pattern Recognition}, pages 6741--6749, 2019.

\bibitem{kolve2017ai2}
Eric Kolve, Roozbeh Mottaghi, Winson Han, Eli VanderBilt, Luca Weihs, Alvaro
  Herrasti, Daniel Gordon, Yuke Zhu, Abhinav Gupta, and Ali Farhadi.
\newblock Ai2-thor: An interactive 3d environment for visual ai.
\newblock {\em arXiv preprint arXiv:1712.05474}, 2017.

\bibitem{krantz2021waypoint}
Jacob Krantz, Aaron Gokaslan, Dhruv Batra, Stefan Lee, and Oleksandr Maksymets.
\newblock Waypoint models for instruction-guided navigation in continuous
  environments.
\newblock In {\em Proceedings of the IEEE/CVF International Conference on
  Computer Vision}, pages 15162--15171, 2021.

\bibitem{krantz2020navgraph}
Jacob Krantz, Erik Wijmans, Arjun Majumdar, Dhruv Batra, and Stefan Lee.
\newblock Beyond the nav-graph: Vision-and-language navigation in continuous
  environments.
\newblock In {\em European Conference on Computer Vision}, 2020.

\bibitem{anderson2020rxr}
Alexander Ku, Peter Anderson, Roma Patel, Eugene Ie, and Jason Baldridge.
\newblock Room-across-room: Multilingual vision-and-language navigation with
  dense spatiotemporal grounding.
\newblock In {\em Proceedings of the 2020 Conference on Empirical Methods in
  Natural Language Processing (EMNLP)}, pages 4392--4412, 2020.

\bibitem{li2019robust}
Xiujun Li, Chunyuan Li, Qiaolin Xia, Yonatan Bisk, Asli Celikyilmaz, Jianfeng
  Gao, Noah~A Smith, and Yejin Choi.
\newblock Robust navigation with language pretraining and stochastic sampling.
\newblock In {\em Proceedings of the 2019 Conference on Empirical Methods in
  Natural Language Processing and the 9th International Joint Conference on
  Natural Language Processing (EMNLP-IJCNLP)}, pages 1494--1499, 2019.

\bibitem{loshchilov2017decoupled}
Ilya Loshchilov and Frank Hutter.
\newblock Decoupled weight decay regularization.
\newblock {\em arXiv preprint arXiv:1711.05101}, 2017.

\bibitem{ma2019self}
Chih-Yao Ma, Jiasen Lu, Zuxuan Wu, Ghassan AlRegib, Zsolt Kira, Richard Socher,
  and Caiming Xiong.
\newblock Self-monitoring navigation agent via auxiliary progress estimation.
\newblock In {\em Proceedings of the International Conference on Learning
  Representations (ICLR)}, 2019.

\bibitem{ma2019regretful}
Chih-Yao Ma, Zuxuan Wu, Ghassan AlRegib, Caiming Xiong, and Zsolt Kira.
\newblock The regretful agent: Heuristic-aided navigation through progress
  estimation.
\newblock In {\em Proceedings of the IEEE Conference on Computer Vision and
  Pattern Recognition}, pages 6732--6740, 2019.

\bibitem{majumdar2020improving}
Arjun Majumdar, Ayush Shrivastava, Stefan Lee, Peter Anderson, Devi Parikh, and
  Dhruv Batra.
\newblock Improving vision-and-language navigation with image-text pairs from
  the web.
\newblock {\em In Proceedings of the European Conference on Computer Vision},
  2020.

\bibitem{meng2020scaling}
Xiangyun Meng, Nathan Ratliff, Yu Xiang, and Dieter Fox.
\newblock Scaling local control to large-scale topological navigation.
\newblock In {\em 2020 IEEE International Conference on Robotics and Automation
  (ICRA)}, pages 672--678. IEEE, 2020.

\bibitem{nguyen2019help}
Khanh Nguyen and Hal Daum{\'e}~III.
\newblock Help, anna! visual navigation with natural multimodal assistance via
  retrospective curiosity-encouraging imitation learning.
\newblock In {\em Proceedings of the 2019 Conference on Empirical Methods in
  Natural Language Processing and the 9th International Joint Conference on
  Natural Language Processing (EMNLP-IJCNLP)}, pages 684--695, 2019.

\bibitem{parvaneh2020counterfactual}
Amin Parvaneh, Ehsan Abbasnejad, Damien Teney, Qinfeng Shi, and Anton van~den
  Hengel.
\newblock Counterfactual vision-and-language navigation: Unravelling the
  unseen.
\newblock {\em Advances in Neural Information Processing Systems}, 33, 2020.

\bibitem{paszke2019pytorch}
Adam Paszke, Sam Gross, Francisco Massa, Adam Lerer, James Bradbury, Gregory
  Chanan, Trevor Killeen, Zeming Lin, Natalia Gimelshein, Luca Antiga, et~al.
\newblock Pytorch: An imperative style, high-performance deep learning library.
\newblock {\em Advances in neural information processing systems},
  32:8026--8037, 2019.

\bibitem{qi2020reverie}
Yuankai Qi, Qi Wu, Peter Anderson, Xin Wang, William~Yang Wang, Chunhua Shen,
  and Anton van~den Hengel.
\newblock Reverie: Remote embodied visual referring expression in real indoor
  environments.
\newblock In {\em Proceedings of the IEEE/CVF Conference on Computer Vision and
  Pattern Recognition}, pages 9982--9991, 2020.

\bibitem{raychaudhuri2021law}
Sonia Raychaudhuri, Saim Wani, Shivansh Patel, Unnat Jain, and Angel Chang.
\newblock Language-aligned waypoint (law) supervision for vision-and-language
  navigation in continuous environments.
\newblock In {\em Proceedings of the 2021 Conference on Empirical Methods in
  Natural Language Processing}, pages 4018--4028, 2021.

\bibitem{ronneberger2015u}
Olaf Ronneberger, Philipp Fischer, and Thomas Brox.
\newblock U-net: Convolutional networks for biomedical image segmentation.
\newblock In {\em International Conference on Medical image computing and
  computer-assisted intervention}, pages 234--241. Springer, 2015.

\bibitem{russakovsky2015imagenet}
Olga Russakovsky, Jia Deng, Hao Su, Jonathan Krause, Sanjeev Satheesh, Sean Ma,
  Zhiheng Huang, Andrej Karpathy, Aditya Khosla, Michael Bernstein, et~al.
\newblock Imagenet large scale visual recognition challenge.
\newblock {\em International journal of computer vision}, 115(3):211--252,
  2015.

\bibitem{savinov2018semi}
Nikolay Savinov, Alexey Dosovitskiy, and Vladlen Koltun.
\newblock Semi-parametric topological memory for navigation.
\newblock In {\em International Conference on Learning Representations}, 2018.

\bibitem{savva2019habitat}
Manolis Savva, Abhishek Kadian, Oleksandr Maksymets, Yili Zhao, Erik Wijmans,
  Bhavana Jain, Julian Straub, Jia Liu, Vladlen Koltun, Jitendra Malik, et~al.
\newblock Habitat: A platform for embodied ai research.
\newblock In {\em Proceedings of the IEEE/CVF International Conference on
  Computer Vision}, pages 9339--9347, 2019.

\bibitem{shen2020igibson}
Bokui Shen, Fei Xia, Chengshu Li, Roberto Mart{\'\i}n-Mart{\'\i}n, Linxi Fan,
  Guanzhi Wang, Shyamal Buch, Claudia D'Arpino, Sanjana Srivastava, Lyne~P
  Tchapmi, et~al.
\newblock igibson, a simulation environment for interactive tasks in large
  realisticscenes.
\newblock {\em arXiv preprint arXiv:2012.02924}, 2020.

\bibitem{shridhar2020alfred}
Mohit Shridhar, Jesse Thomason, Daniel Gordon, Yonatan Bisk, Winson Han,
  Roozbeh Mottaghi, Luke Zettlemoyer, and Dieter Fox.
\newblock Alfred: A benchmark for interpreting grounded instructions for
  everyday tasks.
\newblock In {\em Proceedings of the IEEE/CVF conference on computer vision and
  pattern recognition}, pages 10740--10749, 2020.

\bibitem{tan2019learning}
Hao Tan, Licheng Yu, and Mohit Bansal.
\newblock Learning to navigate unseen environments: Back translation with
  environmental dropout.
\newblock In {\em Proceedings of NAACL-HLT}, pages 2610--2621, 2019.

\bibitem{thomason2020vision}
Jesse Thomason, Michael Murray, Maya Cakmak, and Luke Zettlemoyer.
\newblock Vision-and-dialog navigation.
\newblock In {\em Conference on Robot Learning}, pages 394--406, 2020.

\bibitem{vaswani2017attention}
Ashish Vaswani, Noam Shazeer, Niki Parmar, Jakob Uszkoreit, Llion Jones,
  Aidan~N Gomez, {\L}ukasz Kaiser, and Illia Polosukhin.
\newblock Attention is all you need.
\newblock In {\em Advances in neural information processing systems}, pages
  5998--6008, 2017.

\bibitem{wang2021structured}
Hanqing Wang, Wenguan Wang, Wei Liang, Caiming Xiong, and Jianbing Shen.
\newblock Structured scene memory for vision-language navigation.
\newblock In {\em Proceedings of the IEEE/CVF Conference on Computer Vision and
  Pattern Recognition}, pages 8455--8464, 2021.

\bibitem{wang2019reinforced}
Xin Wang, Qiuyuan Huang, Asli Celikyilmaz, Jianfeng Gao, Dinghan Shen,
  Yuan-Fang Wang, William~Yang Wang, and Lei Zhang.
\newblock Reinforced cross-modal matching and self-supervised imitation
  learning for vision-language navigation.
\newblock In {\em Proceedings of the IEEE Conference on Computer Vision and
  Pattern Recognition}, pages 6629--6638, 2019.

\bibitem{wijmans2019dd}
Erik Wijmans, Abhishek Kadian, Ari Morcos, Stefan Lee, Irfan Essa, Devi Parikh,
  Manolis Savva, and Dhruv Batra.
\newblock Dd-ppo: Learning near-perfect pointgoal navigators from 2.5 billion
  frames.
\newblock In {\em International Conference on Learning Representations}, 2019.

\bibitem{wu2018building}
Yi Wu, Yuxin Wu, Georgia Gkioxari, and Yuandong Tian.
\newblock Building generalizable agents with a realistic and rich 3d
  environment.
\newblock {\em arXiv preprint arXiv:1801.02209}, 2018.

\bibitem{xia2018gibson}
Fei Xia, Amir~R Zamir, Zhiyang He, Alexander Sax, Jitendra Malik, and Silvio
  Savarese.
\newblock Gibson env: Real-world perception for embodied agents.
\newblock In {\em Proceedings of the IEEE Conference on Computer Vision and
  Pattern Recognition}, pages 9068--9079, 2018.

\bibitem{yan2018chalet}
Claudia Yan, Dipendra Misra, Andrew Bennnett, Aaron Walsman, Yonatan Bisk, and
  Yoav Artzi.
\newblock Chalet: Cornell house agent learning environment.
\newblock {\em arXiv preprint arXiv:1801.07357}, 2018.

\bibitem{zhu2021soon}
Fengda Zhu, Xiwen Liang, Yi Zhu, Qizhi Yu, Xiaojun Chang, and Xiaodan Liang.
\newblock Soon: Scenario oriented object navigation with graph-based
  exploration.
\newblock In {\em Proceedings of the IEEE/CVF Conference on Computer Vision and
  Pattern Recognition}, pages 12689--12699, 2021.

\end{thebibliography}
}
\clearpage

\appendix
\section*{Appendices}

\section{Connectivity Graph}
We detail the construction of connectivity graphs in Habitat-Matterport3D environments, as well as the adjustments on the R2R-CE trajectories.

\subsection{Graph Construction (\texorpdfstring{$\boldsymbol{\S}$}~4.2\protect\footnote{Link to Section 4.2 in Main Paper.})}

\paragraph{Projection}
We start with the pre-defined connectivity graphs in MP3D environments, and leverage trajectories in datasets to adjust the position of nodes. For each MP3D scene, the corresponding graph that contains a set of nodes is projected to the same scene in Habitat. Note that, each node that is applied by VLN-CE~\cite{krantz2020navgraph}, for creating the continuous ground-truth paths, is projected to the averaged position of points on the ground-truth paths that are closest to this node. However, such projection only fixes a small portion of invalid nodes (\textit{i.e.} nodes within obstacles) and edges (\textit{i.e.} edges intersects with obstacles). To obtain a fully navigable graph, we design a heuristic to further adjust the position of inaccessible nodes in the environment. 

\paragraph{Criteria}
Four criteria are followed to ensure the quality of the resulting graphs: 
(1) Nodes should not adhere to obstacles. (2) As few as number of nodes should be added for correcting invalid edges. (3) Straightness of the edges should be maintained. (4) Nodes connected with an edge shorter than 0.25 meters should be merged. 

\begin{figure}[t]
  \centering
  \includegraphics[width=\columnwidth]{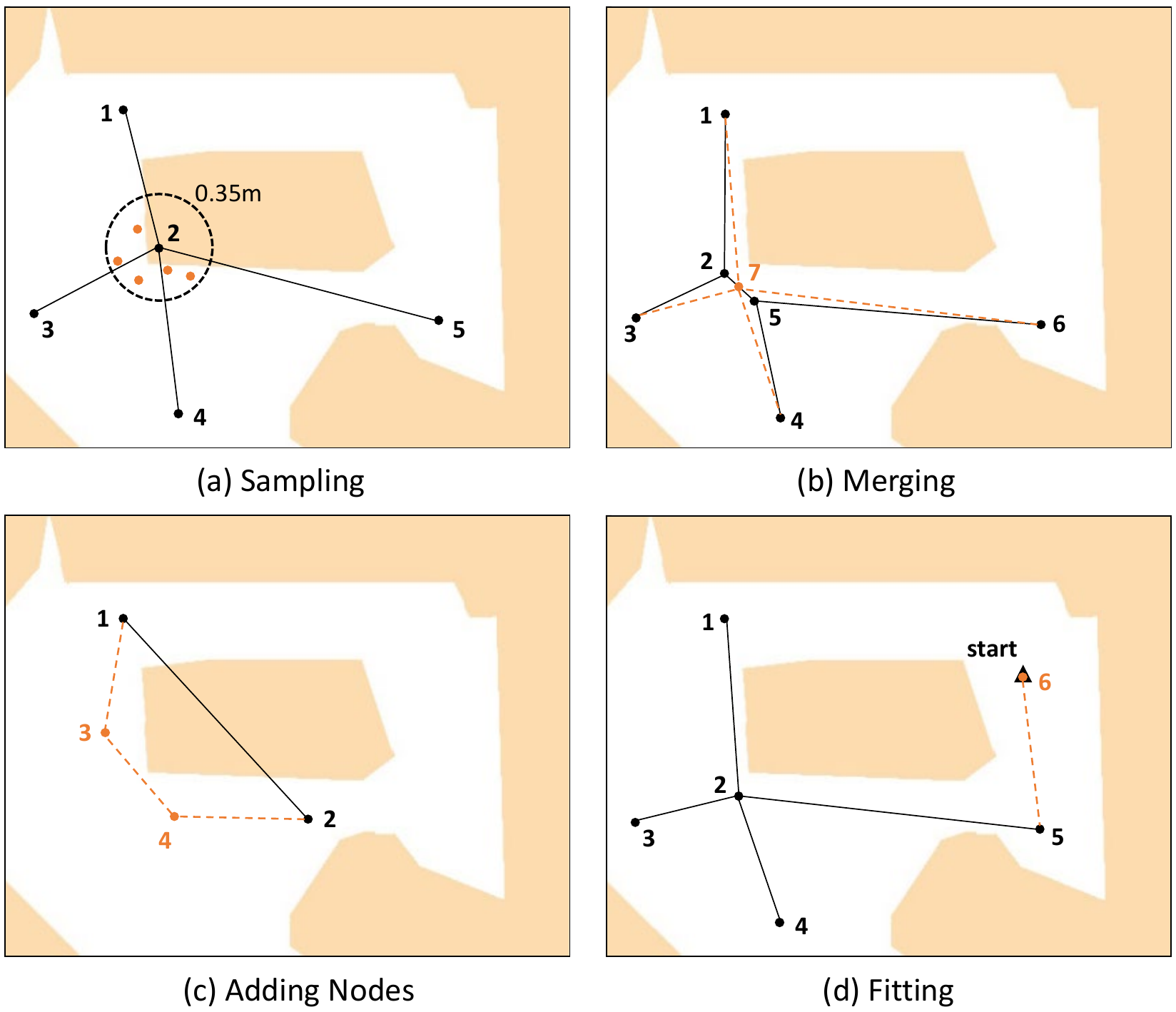}
  \vspace{-15pt}
  \caption{Graph construction methods. (a) The invalid node \textbf{2} is adjusted to a newly sampled position in a 0.35m circle. (b) Node \textbf{2} and node \textbf{5} are merged to a new node \textcolor{orange}{\textbf{7}}, the edges are adjusted accordingly. (c) An invalid edge between nodes \textbf{1} and \textbf{2} is replaced by a detour and two new nodes (\textcolor{orange}{\textbf{3}} and \textcolor{orange}{\textbf{4}}) are added to the graph. (d) New nodes (\textcolor{orange}{\textbf{6}}) are added to the graph if there is no adjacent node to the starting or ending position of a trajectory }
  \label{fig:node_adjust}
\vspace{-5pt}
\end{figure}

\paragraph{Sampling}
For each node that requires an adjustment, we sample 264 points within a 0.35m-radius circle centered at the node as its candidate new positions. Those candidates are uniformly sampled on several rings evenly at \texttt{(0.1,18)}, \texttt{(0.15,30)}, \texttt{(0.2,36)}, \texttt{(0.25,48)}, \texttt{(0.3,60)} and \texttt{(0.35,72)}, where \texttt{(radius,\#samples)}. According to the aforementioned \textit{Criteria}, for all candidate positions that are not within an obstacle, we score them by a weighted sum of three measurements: (1) distance to closest obstacles, (2) number of new nodes that needs to be added, and (3) edge straightness which is computed as the ratio between the sum of lengths of new edges and the original edge length. The candidate position which has the highest score will become the new position of the node.

\paragraph{Adding Nodes}
Detours are sometimes necessary for addressing the invalid edges that go through obstacles, in this case, new nodes are added to the graph. In specific, we apply the \texttt{shortest\_path\_follower} in the \textit{Habitat Simulator}~\cite{savva2019habitat} to generate numbers of paths with various step lengths that connect the two nodes. Nodes within these paths are evaluated using the \textit{Criteria} so that the best set of new nodes can be determined and added to the graph.

\paragraph{Merging}
Once a fully navigable graph is created, nodes within 0.5m are merged to a single node to make the graph more concise. However, this may creates new invalid nodes and edges. As a result, we repeat the \textit{Adjustment}, \textit{Adding Nodes} and \textit{Merging} processes until the entire connectivity graph is fully navigable with edges longer than or equals to 0.5 meters. 

\paragraph{Fitting}
Finally, to match the R2R-CE data~\cite{anderson2018vision,krantz2020navgraph}, we fit the graph for the starting and ending positions of all trajectories in the dataset (except for the \textit{test} split). We check if all the starting and ending points can be matched to nodes on graphs within a geodesic distance of 0.8 meters, if not, new nodes will be added to connect the graph to those positions (using our \textit{Adding Nodes} method).

\begin{figure*}[t]
  \centering
  \includegraphics[width=0.90\textwidth]{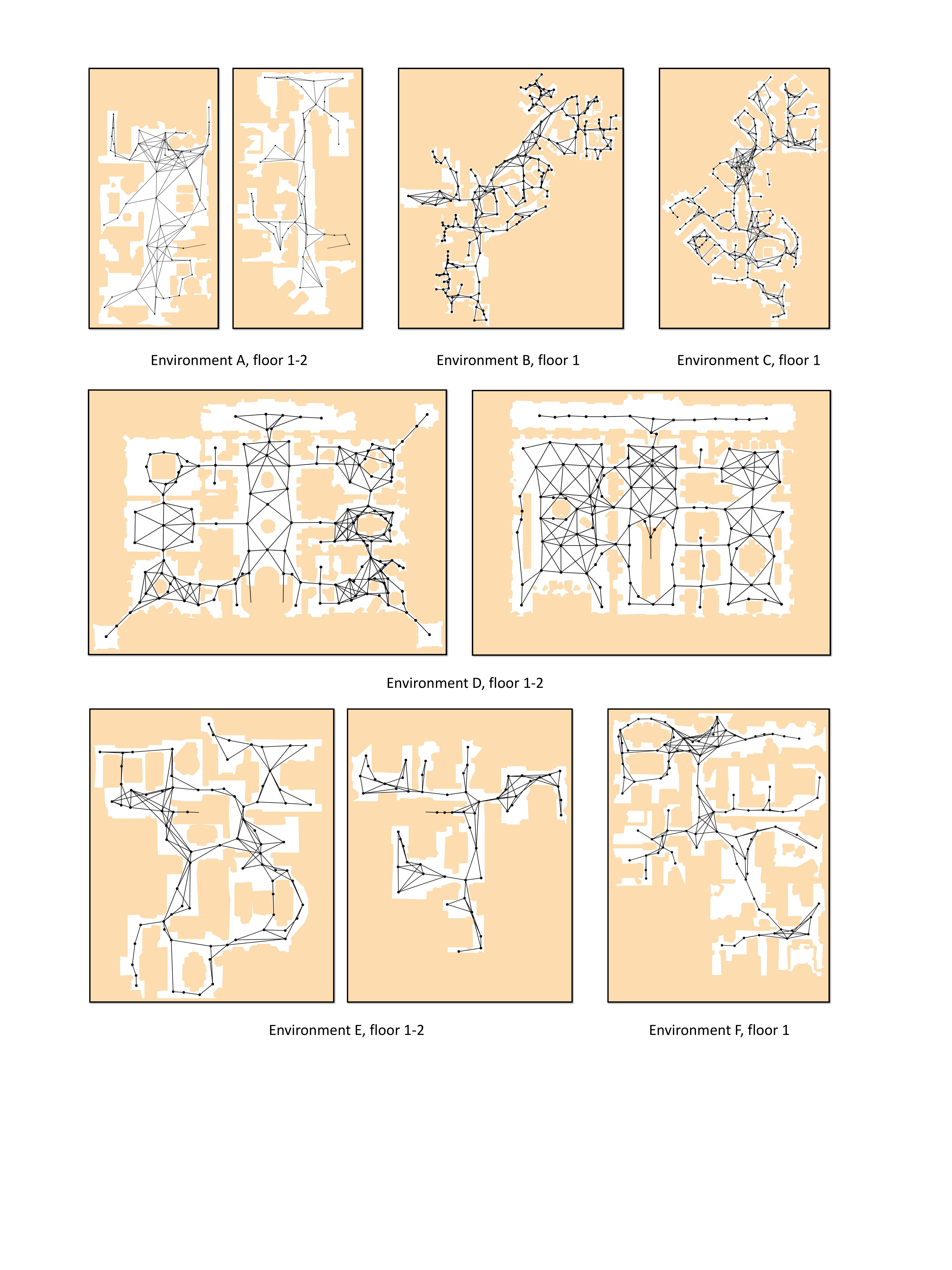}
  \caption{Our Habitat-MP3D graphs.}
  \label{fig:habitat_mp3d_graphs}
\end{figure*}

\begin{figure*}[t]
  \centering
  \includegraphics[width=0.90\textwidth]{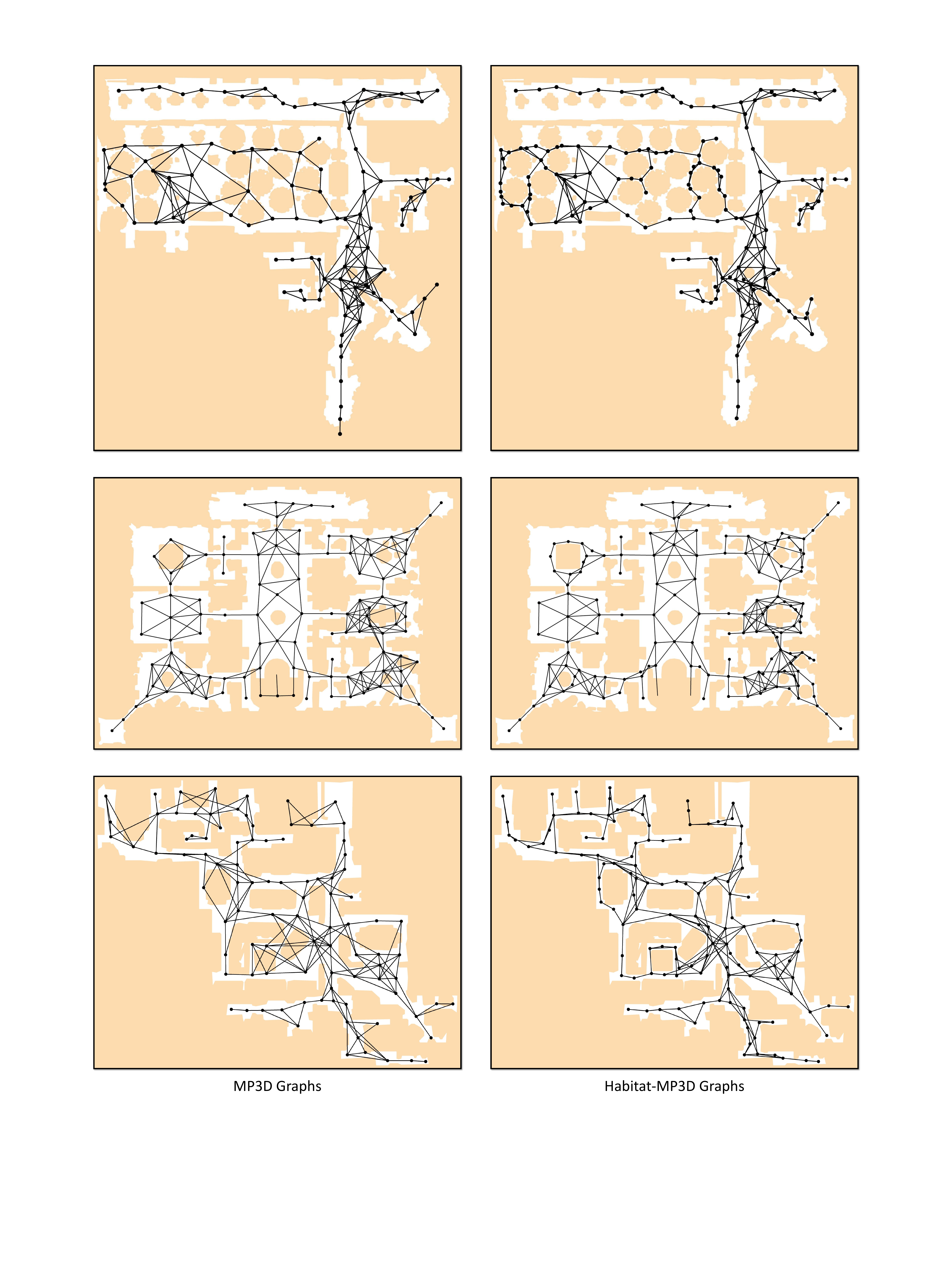}
  \caption{MP3D graphs versus our Habitat-MP3D graphs.}
  \label{fig:mp3d_vs_habitat}
\end{figure*}

\paragraph{Visualization}
Visualizations of the resulting Habitat-MP3D graphs and the comparison to the original MP3D graphs can be found in Figure~\ref{fig:habitat_mp3d_graphs} and Figure~\ref{fig:mp3d_vs_habitat}.

\subsection{R2R-CE Trajectories Adjustment (\texorpdfstring{$\boldsymbol{\S}$}~5.1)}
To establish a fair comparison between discrete and continuous navigation in Habitat (\S5.1), we re-compute and filter the original R2R-CE trajectories based on our Habitat-MP3D graph. Specifically, for each continuous trajectory in R2R-CE~\cite{krantz2020navgraph}, we collect the three closest nodes to its starting and the ending points, respectively. Then, for each one of the nine pairs of starting and ending nodes, we collect up to 200 shortest discrete paths on graph from all the possible paths that connect the two nodes. For the resulting 1,800 paths, we measure the nDTW~\cite{ilharco2019general} between each path and the original continuous trajectory, and select the path with the highest score as the discrete ground-truth path on the Habitat-MP3D graph. Samples which have a discrete trajectory with nDTW lower than 0.92 will be discarded from the dataset. Such expensive and strict filtering process ensures that only samples with well-matched discrete and continuous paths remain in the dataset, which is suitable for us to compare navigation with high-level and low-level actions.

Overall, we obtain 10,755, 1,755 and 745 episodes on the train, validation seen, and validation unseen split, respectively. For reference, the original R2R-CE~\cite{krantz2020navgraph} has 10,819, 1,839 and 778 episodes in the data splits. Besides, the averaged number of nodes on a trajectory is 6.00 and 6.63 for samples in R2R~\cite{anderson2018vision} and in our filtered dataset, respectively.

\subsection{Agent Performance (\texorpdfstring{$\boldsymbol{\S}$}~5.2)}

As shown in Table~\ref{tab:agent_hab_vs_mp3d}, we experiment with the same agents on MP3D graphs and on our Habitat-MP3D graphs. Surprisingly, despite the fact that agents receive lower-quality images and navigate on a more complicated graph in Habitat, similar performance are achieved in unseen environments.

\begin{table}[htp]
    \begin{center}
    \resizebox{\columnwidth}{!}{
    \begin{tabular}{l|c|cccc|cccc}
        \hline \hline
        \multicolumn{1}{c|}{\multirow{2}{*}{Model}} & \multicolumn{1}{c|}{\multirow{2}{*}{Val}} & \multicolumn{4}{c|}{MP3D R2R} & \multicolumn{4}{c}{Habitat-MP3D R2R}\Tstrut\\
        \cline{3-10} & & \multicolumn{1}{c}{TL} &
        \multicolumn{1}{c}{NE$\downarrow$} & \multicolumn{1}{c}{SR$\uparrow$} & \multicolumn{1}{c|}{SPL$\uparrow$} &  
        \multicolumn{1}{c}{TL} & \multicolumn{1}{c}{NE$\downarrow$} & \multicolumn{1}{c}{SR$\uparrow$} & \multicolumn{1}{c}{SPL$\uparrow$}\Tstrut\\
        \hline \hline
        \multicolumn{1}{c|}{\multirow{2}{*}{CMA}} & S & 15.20 & 4.29 & 54.95 & 49.67 & 10.21 & 6.28 & 41.48 & 36.58 \\
        & U & 19.68 & 5.54 & 39.97 & 33.36 & 10.47 & 6.51 & 39.32 & 33.89 \\
        \hline
        \multicolumn{1}{c|}{\multirow{2}{*}{\vlnbert}} & S & 16.62 & 4.00 & 54.65 & 46.58 & 14.21 & 4.63 & 52.21 & 42.53 \\
        & U & 16.75 & 4.59 & 51.13 & 42.92 & 14.34 & 5.22 & 48.89 & 40.36 \\
        \hline \hline
    \end{tabular}}
    \end{center}
    \vspace{-10pt}
    \caption{Performance of agents navigating on MP3D and Habitat-MP3D graphs in Seen (\textit{S}) and Unseen (\textit{U}) environments.}
    \label{tab:agent_hab_vs_mp3d}
    \vspace{-10pt}
\end{table}


\section{Navigator Networks}

In this section, we provide the implementation details of the Cross-Modal Matching agent (CMA)~\cite{wang2019reinforced} and the Recurrent VLN-BERT Agent (\vlnbert)~\cite{hong2020recurrent} on the R2R-CE~\cite{anderson2018vision} and the RxR-CE~\cite{anderson2020rxr} datasets.

\subsection{Architecture (\texorpdfstring{$\boldsymbol{\S}$}~5.1)}

\paragraph{Visual Encoders} Agents in our experiments apply the same RGB and depth encoders to process the candidate images at waypoint directions. As in VLN-CE~\cite{krantz2020navgraph}, we use two ResNet-50~\cite{he2016deep}, one pre-trained on ImageNet~\cite{russakovsky2015imagenet} for classification and another one pre-trained on Gibson~\cite{xia2018gibson} for point-goal navigation~\cite{wijmans2019dd}, to encode the RGB and depth inputs, respectively. These encoders are freezed while training the navigators, where the outputs are fed into the candidate waypoints predictor to infer adjacent waypoints, and into the navigators for \textit{view selection}. 

Refer to the \textit{Main Paper} \S3, we denote the encoded representations as $\{\boldsymbol{v}^{rgb}_1,\boldsymbol{v}^{rgb}_2,\ldots,\boldsymbol{v}^{rgb}_k \mid \boldsymbol{v}^{rgb}_{i} \in \mathbb{R}^{2048}\}$ and $\{\boldsymbol{v}^{d}_1,\boldsymbol{v}^{d}_2,\ldots,\boldsymbol{v}^{d}_k \mid \boldsymbol{v}^{d}_{i} \in \mathbb{R}^{128}\}$\footnote{Subscript $t$ for time step is omitted here for simplicity.}, corresponding to the $k$ directions with waypoints.
The RGB and depth representations are merged before passing to the policy networks as
\begin{equation}
\boldsymbol{f}_{\!i}=\left[\boldsymbol{v}^{rgb}_{i}\boldsymbol{W}_{\!rgb};\boldsymbol{v}^{d}_{i}\boldsymbol{W}_{\!depth};\boldsymbol{d}_{i}\right]\boldsymbol{W}_{\!merge}
\label{eqn:vis_encode}
\end{equation}
where $\boldsymbol{W}$ are learnable linear projections with ReLU activation. Following previous works~\cite{fried2018speaker,tan2019learning}, we explicitly encode the relative direction of each candidate view as $\boldsymbol{d}_{i}$. $\boldsymbol{d}_{i}$ is a vector formed by replicating $(\text{cos}\theta^{i}_{t}, \text{sin}\theta^{i}_{t})$ by 32 times, where $\theta^{i}_{t}$ is the heading angle of the view with respect to the agent's orientation. Note that, unlike previous works, $\boldsymbol{d}_{i}$ in our experiments does not involve an elevation angle because the waypoints predictor only creates a 2D-planar graph. We suggest that predicting 3D waypoints could be a valuable extension for future work. Finally, the overall visual feature $\boldsymbol{f}_{i}$ is of dimension 512 and 768 for the CMA and the \vlnbert~(to match the default hidden dimension of V\&L BERT~\cite{hao2020towards,hong2020recurrent}) models, respectively.

\paragraph{Language Encoders and Initial States}
Given a natural language instruction $\boldsymbol{U}$ of a sequence of $l$ words ${\langle}w_1, w_2, \ldots, w_l{\rangle}$, the agents first encode the instruction into textual representations. For CMA, a bidirectional LSTM~\cite{hochreiter1997long} is applied to encode $\boldsymbol{U}$ with randomly initialized word embeddings as
\begin{equation}
\boldsymbol{X}=\left\langle\boldsymbol{x_1}, \boldsymbol{x_2}, \ldots, \boldsymbol{x_l}\right\rangle=\text{Bi-LSTM}\left(w_1, w_2, \ldots, w_l\right)
\label{eqn:lang_lstm}
\end{equation}
For \vlnbert, the language stream of the two-stream V\&L transformers~\cite{hao2020towards} is applied to process $\boldsymbol{U}$ as
\begin{equation}
\boldsymbol{s}_{0},\boldsymbol{X}=\text{VLN}\CircleArrowright\text{BERT}(\texttt{[CLS]},\boldsymbol{U},\texttt{[SEP]})
\label{eqn:lang_bert}
\end{equation}
with word embeddings pre-trained in PREVALENT~\cite{hao2020towards}. \texttt{[CLS]} and \texttt{[SEP]} are the classification token and the separation token pre-defined in BERT~\cite{devlin2019bert}. \vlnbert~adopts the output of the \texttt{[CLS]} token $\boldsymbol{s}_{0}$ as the agent's initial state, whereas CMA uses a vector of zeros to represent the initial state.

\paragraph{CMA}
We apply the original CMA~\cite{wang2019reinforced} as the policy network in our experiments, which is different from the CMA implemented in VLN-CE~\cite{krantz2020navgraph} that has an additional recurrent module. In specific, at each navigation step, the agent's state is encoded as:
\begin{equation}
\boldsymbol{s}_{t}=\text{GRU}\left(\left[\boldsymbol{c}^{vis,0}_{t};\boldsymbol{a}_{t-1}\boldsymbol{W}_{act}\right],\boldsymbol{s}_{t-1}\right)
\label{eqn:cma_state}
\end{equation}
where $\boldsymbol{a}_{t-1}$ is the agent's past decision represented by the aforementioned directional encoding and $\boldsymbol{W}_{\!act}$ is a learnable projection with a hyperbolic tangent activation. $\boldsymbol{c}^{vis,0}_{t}=\text{SoftATTN}(\boldsymbol{f}_{t}, \boldsymbol{s}_{t-1})$ is the weighted sum of the candidate features $\boldsymbol{f}_{t}$, where the weights are produced by a dot-product based soft-attention \cite{vaswani2017attention} using the agent's past state $\boldsymbol{s}_{t-1}$ as query.
Then, the current state $\boldsymbol{s}_{t}$ is applied to compute the updated textual and visual features as $\boldsymbol{c}^{lang}_{t}=\text{SoftATTN}(\boldsymbol{X}, \boldsymbol{s}_{t})$ and $\boldsymbol{c}^{vis,1}_{t}=\text{SoftATTN}(\boldsymbol{f}_{t}, \boldsymbol{s}_{t})$. Finally, the probability of each candidate directions is computed as
\begin{equation}
p_{t,i}=\text{Softmax}\left(\left[\boldsymbol{s}_{t};\boldsymbol{c}^{vis,1}_{t};\boldsymbol{c}^{lang}_{t}\right]\boldsymbol{W}_{\!a}\left(\boldsymbol{f}_{t,i}\boldsymbol{W}_{\!b}\right)^{T}\right)
\label{eqn:cma_action}
\end{equation}
In inference, the agent will select the view with the greatest probability and navigate to the waypoint in that view.

\paragraph{\vlnbert}
We apply the PREVALENT variant~\cite{hao2020towards} of the \vlnbert~\cite{hong2020recurrent} with slight modifications in our experiments. Specifically, we remove the \textit{cross-modal matching} in \textit{state refinement} since the method complicates the network while leading to a trivial improvement. \vlnbert~applies a multi-layer transformers to perform cross-modal soft-attention across the agent state, encoded language and visual features:
\begin{equation}
\boldsymbol{s}_{t},\boldsymbol{p}_{t}=\text{VLN}\CircleArrowright\text{BERT}(\boldsymbol{s}^{d}_{t-1},\boldsymbol{X},\boldsymbol{f}_{t})
\label{eqn:bert_state}
\end{equation}
where $\boldsymbol{s}^{d}_{t-1}=\left[\boldsymbol{s}_{t-1};\boldsymbol{a}_{t-1}\right]\boldsymbol{\!W}_{act}$ is the previous state with directional encoding, $\boldsymbol{p}_{t}$ is the action probabilities computed as the mean attention weights of the visual tokens $\boldsymbol{f}_{t}$ over all the attention heads in the last transformer layer with respect to the state.

\subsection{Training (\texorpdfstring{$\boldsymbol{\S}$}~5.1)}
All agents in our experiments are trained using imitation learning (IL) with a cross-entropy loss on the action probabilities $p_{t}$ and the oracle action $a^{*}_{t}$ as
$\mathcal{L}_{IL} = - \sum_{t} a^{*}_{t}\text{log}\left(p_{t}\right)$
where the oracle action is to the waypoint which has the shortest geodesic distance to the target, among all predicted waypoints. The oracle stop is positive when the geodesic distance between the agent and the target is shorter than 1.5 meters. Note that, in some rare cases (Figure~\ref{fig:traj_vis_1}), the candidate waypoints predictor might not be able to return a waypoint that brings the agent closer to the target\footnote{We experiment with taking the oracle action at each step for samples in unseen environments in the original VLN-CE data, results show that only 2\% of agents cannot reach within 3 meters to the target.}. In order to allow the agent to explore the environment while learning from teacher actions, we control the agent with schedule sampling~\cite{bengio2015scheduled} to sample an action between oracle and prediction at each step. 

\paragraph{Oracle Actions in RxR-CE}
Each path in RxR-CE~\cite{anderson2020rxr} is composed of multiple shortest sub-paths traversing through a sequence of rooms, which is not necessarily the shortest path to the target. As a result, we design a different method to determine the oracle actions: At each time step, we compute a sub-goal as the intersection of a 3-meter ring centered at the agent and the ground-truth path. Then, the oracle action is to the waypoint which has the shortest sum of geodesic distances from the agent to the waypoint and from the waypoint to the sub-goal. If there is more than one intersection, we use the one that is the farthest on the ground-truth path as the sub-goal. If there is no intersection, we apply the latest sub-goal as the current sub-goal to push the agent to return to the ground-truth path.

\paragraph{Initialization}
The \vlnbert~is initialized from the pre-trained PREVALENT model~\cite{hao2020towards}. When training on RxR-CE~\cite{anderson2020rxr}, we apply the multilingual BERT features to initialize the word embeddings in both the CMA and \vlnbert. We train each model three times each for a different language and combine the results at the end for submitting to the test server.

\paragraph{Hardware and Time Cost} On R2R-CE, we train the CMA and the \vlnbert~for 50 epochs with a batch size of 16, both models take about 3.5 days to complete using a single NVIDIA RTX 3090 GPU. On RxR-CE, we train the models for 25 epochs on a single GPU, with a batch size of 16 and 8 for the CMA and the \vlnbert, respectively. The training takes about 3 days per language due to the multilingual and larger dataset, and longer paths.

\subsection{Evaluation Metrics (\texorpdfstring{$\boldsymbol{\S}$}~3.1)}
Trajectory Length (TL): the average navigation path length in meters, Navigation Error (NE): the average distance between the target and the agent's final position in meters, Success Rate (SR): the ratio of stopping within 3 meters to the target, Success weighted by the normalized inverse of the Path Length (SPL) \cite{anderson2018evaluation}, normalized Dynamic Time Warping (nDTW) and Success weighted by normalized Dynamic Time Warping (SDTW) \cite{ilharco2019general}. While SR focuses on the accuracy of agent's decisions, SPL measures if the agent navigates efficiently, nDTW and SDTW measure if the agent follows the given instructions by computing the similarity between the executed path and the ground-truth path.

\clearpage



\section{Visualization (\texorpdfstring{$\boldsymbol{\S}$}~5.2)}
\begin{figure*}[t]
  \centering
  \includegraphics[width=0.85\textwidth]{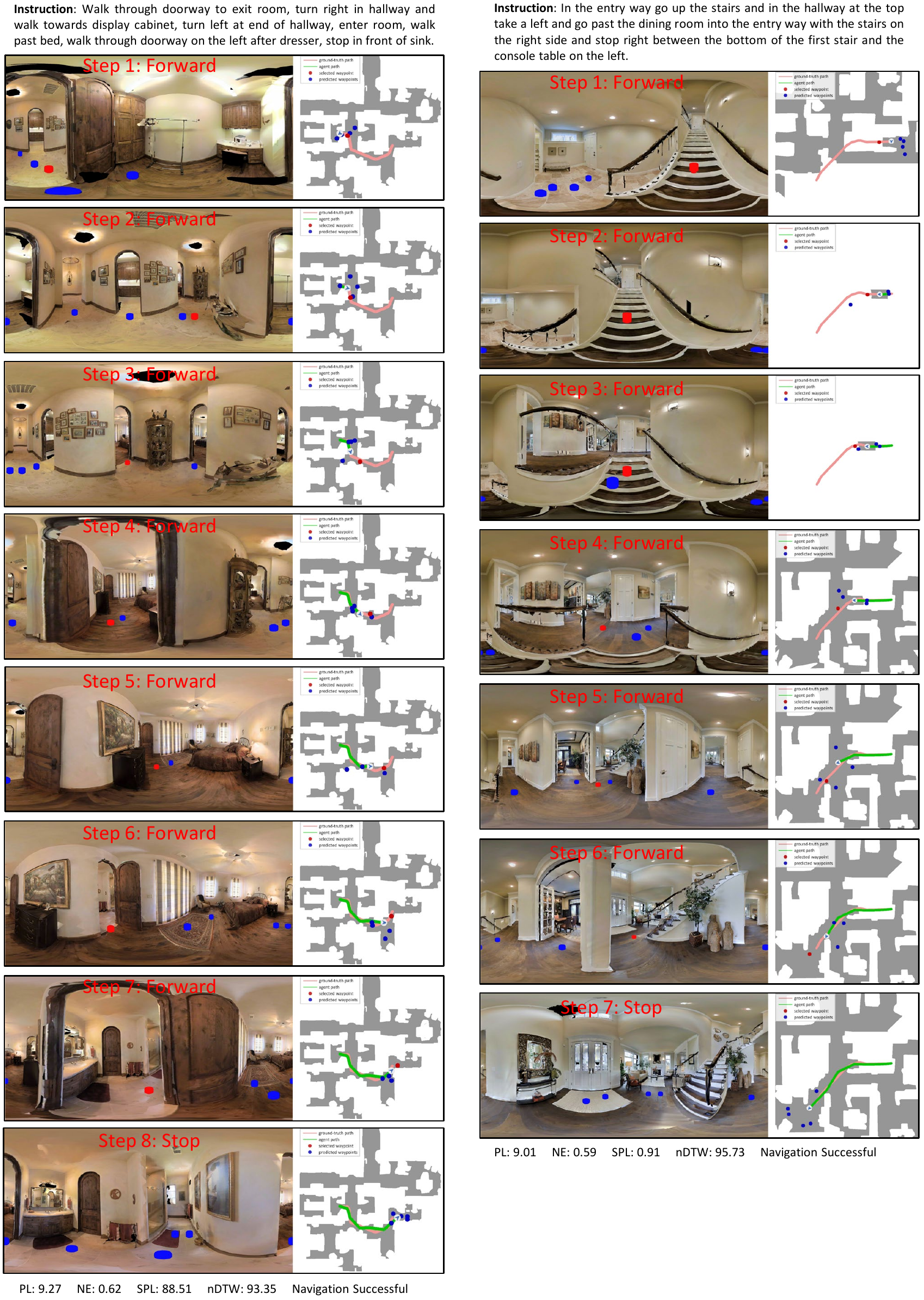}
  \caption{Visualization of trajectories and predicted waypoints. Both samples are successful cases in unseen environments.}
  \label{fig:traj_vis_0}
\end{figure*}

\begin{figure*}[t]
  \centering
  \includegraphics[width=0.85\textwidth]{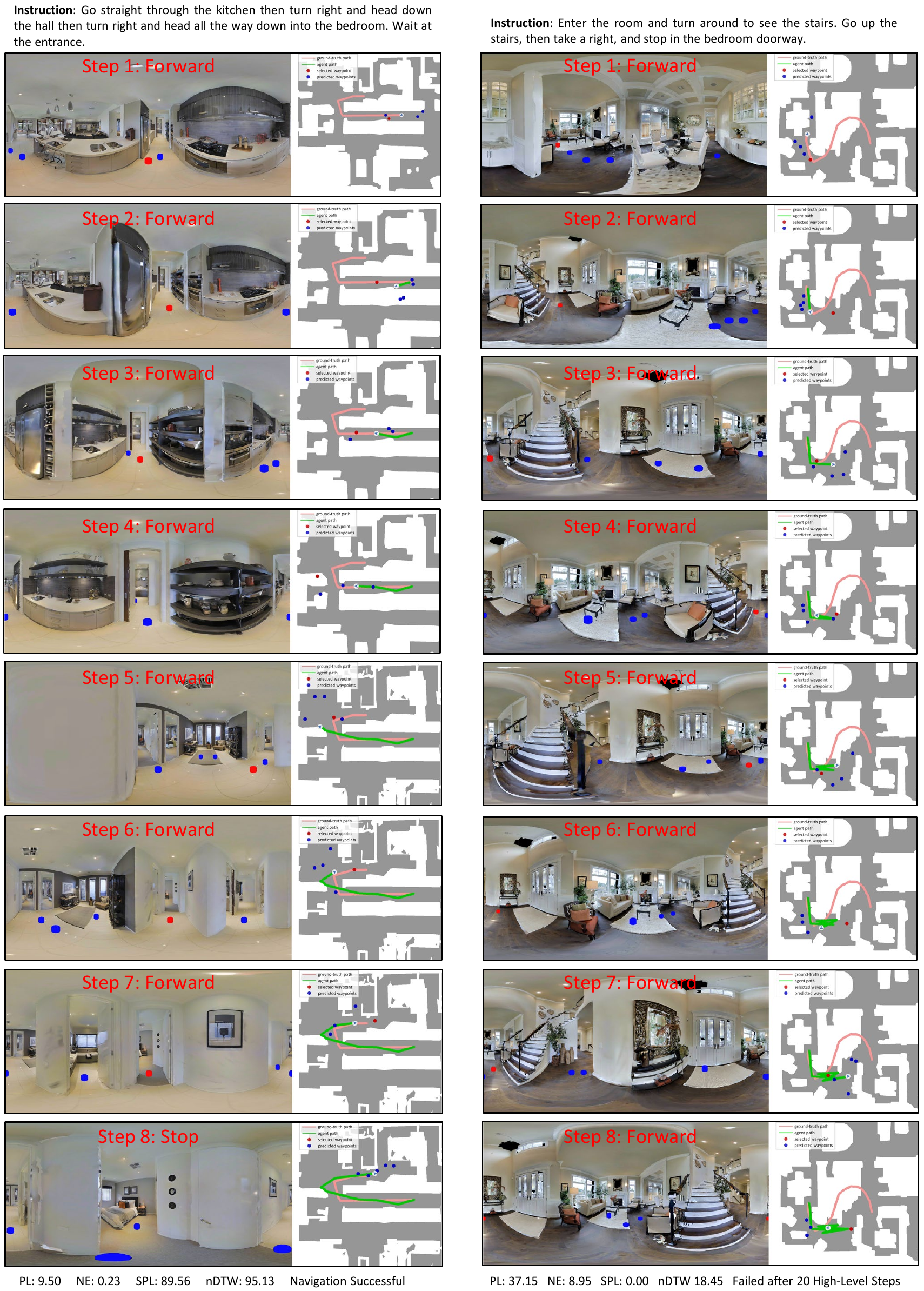}
  \caption{Visualization of trajectories and predicted waypoints. Left: The waypoints predictor produces inaccessible waypoints during navigation (step 2), but the agent avoids choosing those waypoints and eventually reaches the target. Right: The waypoints predictor fails to provide a waypoint on the staircase, so that the agent wanders around the landing and unable to reach the target.}
  \label{fig:traj_vis_1}
\end{figure*}

Figure~\ref{fig:traj_vis_0} and Figure~\ref{fig:traj_vis_1} visualize the agent's trajectories and the predicted waypoints at each step. As shown by the waypoints (indicated with cylinders) in panoramas and the occupancy maps, most of the predictions are nicely positioned at accessible spaces, pointing towards explorable directions around the agent. Thanks to the predicted waypoints, the agent often only needs to make a few decisions for completing a long navigation task. However, in some cases, \textit{e.g.} Figure~\ref{fig:traj_vis_1}.(Right); the agent is not able to explore certain part of the environment when the predictor fails to produce a waypoints, which disturbs the training and inhibits the navigation. We suggest that future work to utilize the online structural information (\textit{e.g.} depth) and agent's behavior, or to make the predictor to collaborate with the agent for producing more flexible waypoints.


\section{Simulator Configurations (\texorpdfstring{$\boldsymbol{\S}$}~5.1)}

According to the official configurations\footnote{R2R-CE code: \href{https://github.com/jacobkrantz/VLN-CE}{https://github.com/jacobkrantz/VLN-CE}, RxR-CE code: \href{https://github.com/jacobkrantz/VLN-CE/tree/rxr-habitat-challenge}{https://github.com/jacobkrantz/VLN-CE/tree/rxr-habitat-challenge}.}, agents in R2R-CE~\cite{anderson2018vision,krantz2020navgraph} and in RxR-CE~\cite{anderson2020rxr} are set up differently in Habitat~\cite{savva2019habitat}. For example, agent in RxR-CE is shorter in height so that it can access places in the environments where an R2R-CE agent cannot reach. For a fair comparison to previous works, agents in our experiments are set up with the standard dimensions for R2R-CE and RxR-CE, respectively. On the other hand, to facilitate the use of the same waypoints predictor and the powerful pre-trained depth-encoder~\cite{wijmans2019dd} in the two datasets, as well as to decrease the rendering cost, we adjust the camera parameters in RxR-CE. Some key configurations are listed here, where the commented numbers are the original values.

\paragraph{R2R-CE}
Configurations: \\

\noindent\fbox{%
    \parbox{\columnwidth}{%
        
        \texttt{~~FORWARD\_STEP\_SIZE: 0.25}
        
        \texttt{~~TURN\_ANGLE: 3x~~~\#30}
        
        \texttt{~~AGENT:}
        
        \texttt{~~~~HEIGHT: 1.50}
        
        \texttt{~~~~RADIUS: 0.10}

        \texttt{~~HABITAT\_SIM:}
        
        \texttt{~~~~ALLOW\_SLIDING: True}
        
        \texttt{~~RGB\_SENSOR:}
        
        \texttt{~~~~WIDTH: 224}
        
        \texttt{~~~~HEIGHT: 224}
        
        \texttt{~~~~HFOV: 90}
        
        \texttt{~~DEPTH\_SENSOR:}

        \texttt{~~~~WIDTH: 256}
        
        \texttt{~~~~HEIGHT: 256}
        
        \texttt{~~~~HFOV: 90}
    }%
}

\paragraph{RxR-CE} Configurations: \\

\noindent\fbox{%
    \parbox{\columnwidth}{%
        
        \texttt{~~FORWARD\_STEP\_SIZE: 0.25}
        
        \texttt{~~TURN\_ANGLE: 3x~~~\#30}
        
        \texttt{~~AGENT:}
        
        \texttt{~~~~HEIGHT: 0.88}
        
        \texttt{~~~~RADIUS: 0.18}
        
        \texttt{~~HABITAT\_SIM:}
        
        \texttt{~~~~ALLOW\_SLIDING: False}
        
        \texttt{~~RGB\_SENSOR:}
        
        \texttt{~~~~WIDTH: 224~~~\#640}
        
        \texttt{~~~~HEIGHT: 224~~~\#480}
        
        \texttt{~~~~HFOV: 90~~~\#79}
        
        \texttt{~~DEPTH\_SENSOR:}

        \texttt{~~~~WIDTH: 256~~~\#640}
        
        \texttt{~~~~HEIGHT: 256~~~\#480}
        
        \texttt{~~~~HFOV: 90~~~\#79}
    }%
}
\\

\noindent\texttt{3x} indicates that the step-wise turning angle is a value in $(0^{\circ}, 3^{\circ}, \ldots, 357^{\circ})$, according to the angular precision of the waypoints predictor.

\paragraph{Turning Angles}
We would like to point out that a \texttt{TURN\_ANGLE} of 30$^{\circ}$ is likely to be unreasonable in the official RxR-CE configurations because such a large turning angle will prevent the agent from heading to certain navigable directions. Moreover, since the ground-truth continuous paths (and the step-wise ground-truth actions) in RxR-CE are generated using a 30$^{\circ}$ turning angle, it results in a large number of zigzag steps while the agent should walks a straight line\footnote{The inflection coefficient of steps is only 1.9 for trajectories in RxR-CE, suggesting that the agent changes actions unreasonably frequently. This number is about 3.2 in R2R-CE when using a 15$^{\circ}$ turning angle.}. As a result, the original ground-truth paths are inappropriate for imitation learning or for evaluating SPL and nDTW. On the other hand, our waypoints predictor enables an efficient imitation learning without using ground-truths computed with a pre-defined fixed angle.

\paragraph{Sliding}
Unlike R2R-CE, agents in RxR-CE are not allowed to slide along obstacles on collision, which drastically increases the difficulty of the task and results in agents that can easily fall into deadlocks. To address this issue, we first constrain the vast majority of predicted waypoints to be positioned in open space by stopping augmenting waypoints during training, so that an agent has less probability of hitting obstacles. Moreover, we implement a simple function which runs some test steps to check for navigability around the agent when it is stuck at the same position, and assists the agent to escape from deadlocks.


\begin{table}[htp]
    \begin{center}
    \resizebox{\columnwidth}{!}{
    \begin{tabular}{l|cccc|cccc}
        \hline \hline
        \multicolumn{1}{c|}{\multirow{2}{*}{Model}} & \multicolumn{4}{c|}{No Sliding Unseen RxR} & \multicolumn{4}{c}{Sliding Unseen RxR}\Tstrut\\
        \cline{2-9} & \multicolumn{1}{c}{NE$\downarrow$} & \multicolumn{1}{c}{SR$\uparrow$} & \multicolumn{1}{c}{SPL$\uparrow$} & \multicolumn{1}{c|}{nDTW$\uparrow$} &  \multicolumn{1}{c}{NE$\downarrow$} & \multicolumn{1}{c}{SR$\uparrow$} & \multicolumn{1}{c}{SPL$\uparrow$} & \multicolumn{1}{c}{nDTW$\uparrow$}\Tstrut\\
        \hline \hline
        \multicolumn{1}{c|}{\multirow{1}{*}{CMA}} & 8.76 & 26.59 & 22.16 & 47.05 & 8.08 & 31.36 & 26.93 & 47.36 \\
        \hline
        \multicolumn{1}{c|}{\multirow{1}{*}{\vlnbert}} & 8.98 & 27.08 & 22.65 & 46.71 & 7.62 & 33.25 & 28.45 & 47.57 \\
        \hline \hline
    \end{tabular}}
    \end{center}
    \vspace{-10pt}
    \caption{Agents performance with and without sliding.}
    \label{tab:agent_sliding}
\end{table}

Table~\ref{tab:agent_sliding} compares the navigation performance in RxR-CE with and without allowing the agents to slide along obstacles on collision. For each model, we train three monolingual agents independently, the results are averaged and presented in the table. By enabling sliding (in which case the waypoint augmentation can be applied), both the CMA and the \vlnbert~achieve significantly better results.


\section{Limitations (\texorpdfstring{$\boldsymbol{\S}$}~4.1 \& \texorpdfstring{$\boldsymbol{\S}$}~5.2)}
In this section, we would like to share some limitations of our proposed method, \textit{i.e.} the candidate waypoints predictor. Although our experiments demonstrate the high accuracy and the strong generalization ability of the module, we believe this information will shine some light on potential room for improvement and greatly benefit future research following this work.

\vspace{-5pt}

\paragraph{Number of Candidates}
We limit the maximum number of predicted waypoints to be 5 at any position to reduce the computational cost and stabilize the training of the agents. However, there exists some spatial structures where a position can lead to more navigable directions. We suggest future work to consider a dynamic number of waypoints at different positions.

\vspace{-5pt}

\paragraph{Prediction at Rare Structures} 
As shown in Figure~\ref{fig:traj_vis_1}, at some rare structures in the environment such as stairs, the candidate waypoints predictor fails to produce a waypoint on stairs and therefore inhibits agent's navigation. The main reason behind this issue is the amount of training samples for the predictor is insufficient at those structures. Future work can identify rare structures to sample more data and improve the loss function to balance the learning.

\vspace{-5pt}

\paragraph{Online Prediction Adjustment}
Our agent fully trusts and applies the predicted waypoints to navigate, which results in issues mentioned above. This problem is more severe in RxR-CE as the agents can easily enter deadlocks but the predicted waypoints cannot help the agents to escape from them. We suggest future work to equip the agent with the ability to adjust waypoints according to the local structure or the agent's control outcomes.

\vspace{-5pt}

\paragraph{Update Waypoints Predictor}
Although our candidate waypoints predictor demonstrates high accuracy in unseen MP3D~\cite{chang2017matterport3d} environments, it is unclear if it can be transferred to other distinct and out-of-domain scenes. It would be valuable to develop a waypoints predictor which can be updated to adapt new environments (even without the pre-defined connectivity graphs).

\vspace{-5pt}

\paragraph{State-Conditioned Waypoints Prediction}
In this work, we argue that decoupling the waypoint prediction and agent's decision making can reduce agent's state space and facilitates learning. However, we also recognize that the state information such as the navigation progress and landmarks in instruction could benefit the waypoints predictor for creating more effective waypoints to reach the target. Future work could combine the two ideas to build a waypoints predictor that promotes the navigation.


\section{Societal Impact}
In this research, we apply the R2R-CE~\cite{anderson2018vision} and the RxR-CE~\cite{anderson2020rxr} datasets, which are available under license from Matterport3D~\cite{chang2017matterport3d}. The datasets contain thousands of photos of indoor environments and instructions in different languages for traversing the environments; None of the photos contain recognizable individuals and none of the instructions contain inappropriate language. Experiments are safely and confidentially performed in the Habitat Simulator~\cite{savva2019habitat}. This research is at the early stages of pushing towards embodied AI that follows human instructions, there are minimal ethical, privacy or safety concerns. In the future, if a such robot is implemented in real-world, it could benefit the society by assisting people with daily work.


\clearpage

\end{document}